\documentclass[letterpaper]{article} 
\usepackage{aaai25}  
\usepackage{times}  
\usepackage{helvet}  
\usepackage{courier}  
\usepackage[hyphens]{url}  
\usepackage{graphicx} 
\usepackage{natbib}  
\usepackage{caption} 
\usepackage{algorithm}
\usepackage{algorithmic}

\usepackage{newfloat}
\usepackage{listings}
\DeclareCaptionStyle{ruled}{labelfont=normalfont,labelsep=colon,strut=off} 
\floatstyle{ruled}
\newfloat{listing}{tb}{lst}{}
\floatname{listing}{Listing}
\usepackage{booktabs}       
\usepackage{amsfonts}       
\usepackage{nicefrac}       
\usepackage{microtype}      
\usepackage{xcolor}         

\usepackage{amsmath}
\usepackage{bm}

\usepackage{graphicx,verbatimbox}
\usepackage{multirow}
\usepackage{pifont} 

\title{Backdoor Attacks against No-Reference Image Quality Assessment Models\\ via a Scalable Trigger}
\author{
  Yi Yu\textsuperscript{\rm 1}, Song Xia\textsuperscript{\rm 1}, Xun Lin\textsuperscript{\rm 2}, Wenhan Yang\textsuperscript{\rm 3}\thanks{Corresponding author}, Shijian Lu\textsuperscript{\rm 1}, Yap-Peng Tan\textsuperscript{\rm 1}, Alex Kot\textsuperscript{\rm 1}
}
\affiliations{
    \textsuperscript{\rm 1}Nanyang Technological University, Singapore\\
    \textsuperscript{\rm 2}Beihang University, Beijing, China\\
    \textsuperscript{\rm 3}Pengcheng Laboratory, Shenzhen, China
    \\
    \{yuyi0010, xias0002\}@e.ntu.edu.sg, linxun@buaa.edu.cn, yangwh@pcl.ac.cn, \{shijian.Lu, eyptan, eackot\}@ntu.edu.sg
}

\usepackage{bibentry}
\begin{document}

\maketitle

\begin{abstract}
No-Reference Image Quality Assessment (NR-IQA), responsible for {assessing the quality of a single input image without using any reference}, plays a critical role in evaluating and optimizing computer vision systems, \textit{e.g.,} low-light enhancement. 
Recent research indicates that NR-IQA models are susceptible to adversarial attacks, which can significantly alter predicted scores with {visually imperceptible} perturbations. 
Despite revealing vulnerabilities, these attack methods have limitations, including high computational demands, untargeted manipulation, limited practical utility in white-box scenarios, and reduced effectiveness in black-box scenarios.
To address these challenges, {we shift our focus to another significant threat} and present a novel poisoning-based backdoor attack against NR-IQA (BAIQA), allowing the attacker to manipulate the IQA model's output to any desired target value by {simply} adjusting {a} scaling coefficient $\alpha$ for the trigger.
We propose to inject the trigger in the discrete cosine transform (DCT) domain to {improve the local invariance of the trigger} for countering trigger diminishment in NR-IQA models {due to widely adopted data augmentations}.
Furthermore, {the universal adversarial perturbations (UAP) in the DCT space are designed as the trigger}, to increase IQA model susceptibility to manipulation and {improve} attack effectiveness.
In addition to the heuristic method for poison-label BAIQA (P-BAIQA), we explore the design of clean-label BAIQA (C-BAIQA), focusing on $\alpha$ sampling and image data refinement, driven by theoretical insights {we reveal}.
%
%
Extensive experiments on diverse datasets and various NR-IQA models demonstrate the effectiveness of our attacks.
\end{abstract}

\begin{links}
    \link{Code and appendix}{https://github.com/yuyi-sd/BAIQA}
\end{links}

\section{Introduction}
\label{intro}
Recently, deep neural networks (DNNs) have achieved superior performance in computer vision~\cite{he2016deep}, including Image Quality Assessment (IQA)~\cite{zhang2021uncertainty}. 
IQA aims to predict image quality {in line with} human perception, categorized into Full-Reference (FR-IQA) and No-Reference (NR-IQA) models based on access to reference images.
While FR-IQA techniques, such as SSIM~\cite{ssim}, LPIPS~\cite{lpips}, and DISTS~\cite{dists}, compare {signal distortions} with reference images, NR-IQA models {aims to simulate} human perceptual judgment to assess the quality of a image without a reference image.
{NR-IQA is adopted in a wide range of applications as evaluation criterion} such as image transport systems~\cite{fu2023asymmetric}, video compression~\cite{rippel2019learned}, and image restoration ~\cite{zhang2019ranksrgan}, highlighting its pivotal role in real-world image processing algorithms. 
Leveraging the capabilities of DNNs, recent NR-IQA models~\cite{yang2022maniqa} have achieved remarkable consistency with humans.

Alongside the impressive performance of DNNs, concerns about their security have grown~\cite{liang2021uncovering,Yu_2022_CVPR,yu2024purify,xiatransferable,xia2024mitigating,wang2024eclipse,wang2024unlearnable}.
Adversarial attacks (AA) on NR-IQA models have recently received considerable attention. These attacks aim to substantially alter predicted scores with {imperceptible} perturbations applied to input images. 
Despite the insights provided by these attacks in NR-IQA models, they exhibit several intrinsic shortcomings:
1) Some attacks~\cite{zhang2022perceptual} assume a white-box scenario, wherein the attacker has full access to the model and its parameters, limiting their practical utility in {the} real world.
2) Certain attacks~\cite{korhonen2022adversarial}, relying on surrogate models to generate adversarial examples and transfer them to the target model, often suffer from reduced effectiveness in {limited} transferability.
3) The generation of adversarial examples is formulated as an optimization problem, {requiring} substantial computational resources and time to iteratively optimize solutions for each sample. 
4) Most attacks~\cite{zhang2022perceptual,zhang2024vulnerabilities} aim to create untargeted attacks, focusing on inducing significant deviations rather than shifting predictions to a specific target.

\begin{figure*}[t]
\centering
\includegraphics[width=0.96\linewidth]{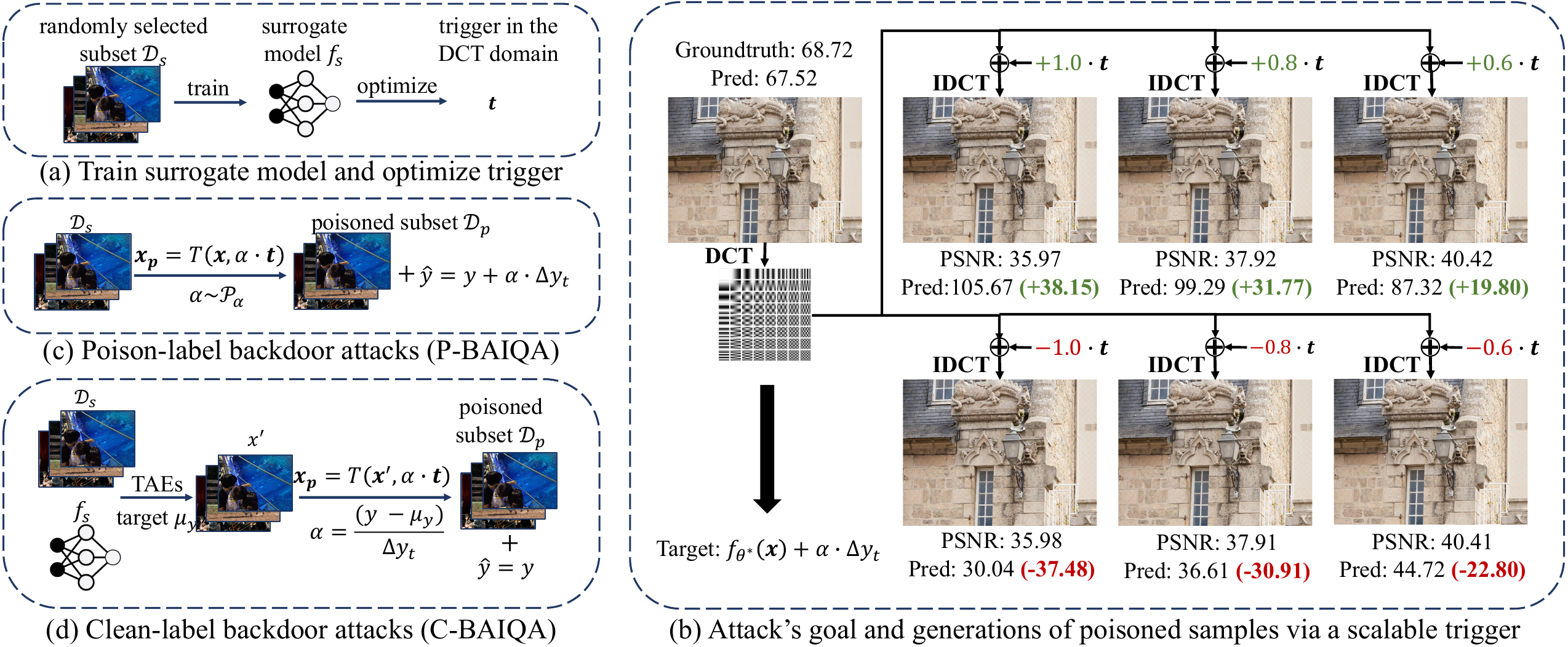}
\caption{
\textbf{1) Poison subset:}
After using (a) to get the trigger $\boldsymbol{t}$, we utilize the trigger injection \( T(\boldsymbol{x}, {\alpha} \!\cdot\! \boldsymbol{t}) \) outlined in (b), enabling the P-BAIQA/C-BAIQA in (c)/(d).  
\textbf{2) Train model:} $f_{\boldsymbol{\theta^{*}}}$ are trained on the set $\mathcal{D}_t$ consisting of a clean subset $\mathcal{D}_c$ and a poisoned subset $\mathcal{D}_p$.
\textbf{3) Attack at test-time:} As shown in (b), attackers can adjust the output to any desired value using $\alpha$ to generate the triggered image \( \boldsymbol{x_p}\!=\!T(\boldsymbol{x}, {\alpha} \!\cdot\! \boldsymbol{t}) \). We offer the PSNR between the clean $\boldsymbol{x}$ and $\boldsymbol{x_p}$, along with the predictions. 
We set $\Delta{y_t}\!=\!40$.
}
\label{teaser}
\end{figure*}

To tackle these challenges, we consider an alternative threat: backdoor attacks (BA), which are more efficient and require no white-box access to the model .
In this scenario, the adversary can inject a stealthy backdoor into the target model corresponding to a unique trigger during training~\cite{saha2020hidden,fang2022backdoor,yu2023backdoor,yu2024chronic,yu2024robust,liu2024backdoor,zheng2024towards}, enabling the compromised model to function normally on benign samples but exhibit malicious behavior when presented with samples containing the trigger during inference.
Data poisoning is the prevailing method for executing BA~\cite{chen2017targeted}, whereby the adversary compromises specific training samples to {embed} the backdoor.
Models trained on poisoned data {with} the trigger, consistently {predicted} a specific target whenever the trigger is present during testing. These attacks are challenging to detect, as the backdoored models maintain high performance on clean {test} data. 
%
%
Current poisoning-based BA {falls} into two categories: 1) poison-label attacks, which contaminate training examples {and} alter their labels to the target value; and 2) clean-label attacks, which solely affect training examples while maintaining their labels.

In this paper, we introduce a novel poisoning-based backdoor attack against NR-IQA via a scalable trigger, addressing both poison-label and clean-label scenarios.
The overall pipeline is illustrated in Fig.~\ref{teaser}.
Unlike traditional classifier BA, which {operates} in discrete outputs, our approach{, designed} for IQA models with continuous outputs, involves embedding a backdoor that manipulates the model's output to any {desired} value using a scaling coefficient $\alpha$ for the trigger as shown in Fig.~\ref{teaser} (b).
To counteract the diminishing effect of triggers in NR-IQA models due to cropping or data augmentation, {we propose global-wise triggers on the patch-based pattern, leveraging the DCT space injections.}
%
Inspired by the effectiveness of universal adversarial perturbations (UAP) in misleading well-trained models, we utilize UAP in the DCT space (UAP-DCT) as triggers as shown in Fig.~\ref{teaser} (a), enhancing the susceptibility of IQA models to manipulations and improving attack effectiveness.
Drawing from our proposed trigger injection and backdoor objectives, we devise an efficient heuristic technique for poison-label attacks (P-BAIQA). 
Furthermore, as shown in Fig.~\ref{teaser} (d), we explore the design of clean-label attacks (C-BAIQA), focusing on the sampling strategy of the scaling coefficient $\alpha$ and the refinement of image data $\boldsymbol{x}$, driven by theoretical insights.

Our main contributions are summarized below:

$\bullet$ We introduce a novel poisoning-based backdoor attack against NR-IQA (BAIQA) via scalable triggers.
{As far as we know, BAIQA is the first method to manipulate the predicted score to any desired value by varying the coefficient $\alpha$}.

$\bullet$ We propose to inject the trigger in the DCT domain to address trigger diminishment due to cropping or data augmentation. 
Furthermore, we utilize UAP in the DCT space as the trigger, to enhance attack effectiveness.

$\bullet$ In addition to the heuristic method for poison-label attacks (P-BAIQA), we explore the {the more challenging} clean-label attacks (C-BAIQA) {with theoretical insights}, focusing on $\alpha$ sampling and image refinement. 
We demonstrate the effectiveness of our attacks on diverse datasets and models, and also show the resistance to several backdoor defenses.

\section{Related Work}
\textbf{Image Quality Assessment.}
IQA tasks {aim} at predicting image quality scores that are consistent with human perception, often represented {by} Mean Opinion Score (MOS). 
These tasks can be categorized into Full-Reference (FR) and No-Reference (NR). 
In FR-IQA, the objective is to predict the quality score of the distorted image by comparing with its reference image. 
However, obtaining reference images can be challenging in real-world scenarios, leading to the emergence of NR-IQA, which predicts using only the distorted image.
Some approaches~\cite{mittal2012no,ghadiyaram2017perceptual} leverage hand-crafted features, while others explore the influence of semantic information. 
For instance, Hyper-IQA~\cite{su2020blindly} employs a hypernetwork to obtain different quality estimators for images with varying content. DBCNN~\cite{zhang2020blind} utilizes two independent neural networks to extract distorted and semantic information from images, which are then combined using bilinear pooling. 
Additionally, various studies have investigated the effectiveness of architectures in NR-IQA. TReS~\cite{golestaneh2021no} and MUSIQ~\cite{ke2021musiq} leverage vision transformers~\cite{dosovitskiy2020image} and demonstrate their efficacy in NR-IQA tasks. 
While adversarial attacks against IQA tasks have received some attention~\cite{zhang2022perceptual, shumitskaya2022universal, liu2024defense,zhang2024vulnerabilities}, a more concerning and practical threat, known as backdoor attacks, remains unexplored.

\noindent\textbf{Backdoor Attacks.}
BA~\cite{chen2017targeted} and AA~\cite{intriguing} intend to modify the benign samples to mislead the DNNs, but they have some intrinsic differences. 
At the inference stage, AA~\cite{DBLP:conf/iclr/MadryMSTV18,ilyas2018black} require much computational resources and time to generate the perturbation through iterative optimizations, and thus are not efficient in deployment.
However, the perturbation (trigger) is known or easy to generate for BA. 
From the perspective of the attacker's capacity, BA have access to poisoning training data, which adds an attacker-specified trigger (\textit{e.g.} a local patch) and alters the corresponding label. 
BA on DNNs have been explored in BadNet~\cite{gu2017badnets} for image classification by poisoning some training samples, and the essential characteristic consists of 1) backdoor stealthiness, 2) attack effectiveness on poisoned images, 3) low performance impact on clean images. 
Based on the capacity of attackers, BA can be categorized into poisoning-based and non-poisoning-based attacks~\cite{li2020backdoor}.
For poisoning-based attacks~\cite{li2020invisible,li2021invisible}, attackers can only manipulate the dataset by inserting poisoned data, and have no access to the model training process. 
In contrast, non-poisoning-based attacks~\cite{dumford2020backdooring,rakin2020tbt,guo2020trojannet,doan2021lira} inject the backdoor by modifying the model parameters instead.
For trigger generations, most attacks~\cite{chen2017targeted,steinhardt2017certified} rely on fixed triggers, and several recent methods~\cite{nguyen2020input,liu2020reflection} extend it to be sample-specific. 
For the trigger domain, several works~\cite{zeng2021rethinking,wang2021backdoor, yue2022invisible} consider the trigger in the frequency domain due to its advantages~\cite{cheng2023frequency}. 
FTrojan~\cite{wang2021backdoor} blockifies images and adds the trigger in the DCT domain, but it uses two fixed channels with fixed magnitudes. 

\section{Methodology}
\subsection{Problem Formulation}
In the context of NR-IQA, consider a trainer to learn a model \( f_{\boldsymbol{\theta}}: \mathcal{X} \!\rightarrow\! \mathcal{Y} \). 
Here, \( \mathcal{X} \!\subseteq\! \mathbb{R}^{H \times W \times 3} \) is the RGB space of inputs,
\( \mathcal{Y} \!\subseteq\! \mathbb{R}\) is the space of outputs,
and $\bm{\theta}$ is the trainable parameters. $H$ and $W$ are the height and width of inputs, respectively.
The learning process of \( f_{\boldsymbol{\theta}}\) involves training with a dataset $\mathcal{D}_{t}$.
\( \mathcal{D}_{t} \) can comprise both clean and poisoned data, denoted as \( \mathcal{D}_{t} = \mathcal{D}_{c} \cup \mathcal{D}_{p} \). 
\( \mathcal{D}_{c} \) consists of correctly labeled clean data, while \( \mathcal{D}_{p} \)
{contains the poisoned data embedded with triggers by the attacker.}
The attacker has two primary objectives: \textbf{1)}
maintaining \textbf{the model's prediction accuracy on clean data}, thereby preventing detection through diminished performance on a validation set.
\textbf{2)} implanting a backdoor mechanism in the IQA model, enabling the attacker to \textbf{manipulate the output by adding triggers} to the input.

\noindent\textbf{Our Backdoor’s Goals.} Unlike traditional BA on classifiers, where the output space is discrete, the output space of IQA is continuous. Hence, our goal is to embed a backdoor capable of manipulating the model to output any values, rather than only one target class typically seen in BA on classifiers. 
This ensures that any input \( \boldsymbol{x} \!\in\! \mathcal{X} \) with MOS value \( y \), once altered with the trigger \( \boldsymbol{t} \) into the poisoned data \( T(\boldsymbol{x}, {\alpha} \!\cdot\! \boldsymbol{t}) \), is incorrectly predicted into a specific target value \( y \!+\! \alpha \!\cdot\! {\Delta{y}}_{t}\). 
Here, $\alpha \!\!\in\!\! [-1,1]$ controls the variations, and regulates the magnitude of $\boldsymbol{t}$. $T$ is the trigger injector, and $\alpha \cdot \Delta{y}_{t}$ is the target variation. 
We can see that when using the Taylor expansion and considering the first-order derivative, the output of $f_{\boldsymbol{\theta}}$ has a certain degree of linearity with respect to $\alpha$, as follows:
\begin{equation}\small
\begin{split}
f_{\boldsymbol{\theta}}(T(\boldsymbol{x}, {\alpha} \cdot \boldsymbol{t})) = f_{\boldsymbol{\theta}}(T(\boldsymbol{x}, 0\cdot \boldsymbol{t})) + \alpha\frac{\partial f_{\boldsymbol{\theta}}(T(\boldsymbol{x}, \alpha_0 \cdot \boldsymbol{t}))}{\partial \alpha_0}\Big|_{\alpha_0=0}.\nonumber
\end{split}
\end{equation}
Thus, by choosing a proper $\alpha$, attackers can manipulate the output to any target.
The overall optimization is given by:
\begin{equation}\small
\begin{split}
& \forall \alpha \in [-1,1]: \max_{{T}(\cdot, {\alpha}\cdot \boldsymbol{t})} \underbrace{\mathop\mathbb{E}\limits_{(\boldsymbol{x},y) \sim \mathcal{C}}\Big[\big\lvert f_{\boldsymbol{\theta}^{*}}(\boldsymbol{x})-y \big\lvert\Big]}_{\text{low impact on clean data}}\\
&\quad\quad\quad\quad+ \underbrace{\mathop\mathbb{E}\limits_{(\boldsymbol{x},y) \sim \mathcal{C}}\Big[\big\lvert f_{\boldsymbol{\theta}^{*}}( T(\boldsymbol{x}, {\alpha} \cdot \boldsymbol{t}))-(y + \alpha \cdot {\Delta{y}}_{t})\big\lvert\Big]}_{\text{backdoor effectiveness}},\\
& ~~\text{s.t.} ~~\boldsymbol{\theta}^{*} = \arg\min_{\boldsymbol{\theta}} \sum_{(\boldsymbol{x}_i, y_i) \in \mathcal{D}_{c}} \mathcal{L}(f_{\boldsymbol{\theta}}(\boldsymbol{x}_i), y_i)\\[-3pt]
&\quad\quad\quad\quad\quad\quad+\sum_{(T(\boldsymbol{x}_i, {\alpha} \cdot \boldsymbol{t}), \hat{y}_i) \in \mathcal{D}_{p}} \mathcal{L}(f_{\boldsymbol{\theta}}(T(\boldsymbol{x}_i, {\alpha} \cdot \boldsymbol{t})), \hat{y}_i),
\end{split}
\label{eq:backdoor_attack}
\end{equation}
where \( \mathcal{C} \) is the distribution of clean data. The loss function \( \mathcal{L} \), the model \(f_{\boldsymbol{\theta}} \), and other hyperparameters are chosen exclusively by trainers. Specifically, when $\alpha=0$, $T(\boldsymbol{x}, {\alpha} \cdot \boldsymbol{t})=\boldsymbol{x}$.

To achieve these objectives, the attacker modifies the benign samples into poisoned ones and constructs a poisoned subset \( \mathcal{D}_{p}\!\!=\!\!\{\!(T(\boldsymbol{x}_i, {\alpha} \cdot \boldsymbol{t}), \hat{y}_i)\!\}\!_{i=1}^{N_p} \). 
To enhance the stealthiness of the attacks, 
{it is common practice to adopt a low poisoning ratio, represented by \(r = {|\mathcal{D}_{p}|}/{|\mathcal{D}_{t}|}\), often below a certain threshold like 20\%.}
Furthermore, if the original label \({y}_i\) in  \(\mathcal{D}_{p}\) are altered to the target value \(y_i + \alpha \cdot {\Delta{y}}_{t}\), the strategy is named \textbf{poison-label attacks}. Otherwise, if the original label is maintained, the strategy is considered \textbf{clean-label attacks}.
The assumptions for our attacks are:
1) {The attacker has no access to the training process, including a lack of knowledge on the model's architecture, and other configurations.}
2) {The attacker can inject poisoned data into the training set, achieved by converting clean data into poisoned data.}

\subsection{DCT Space for Robust Trigger Injection}
\noindent\textbf{Patch-based Patterns.} Since NR-IQA models typically randomly crop multiple patches from input images and compute the average score based on these patches~\cite{su2020blindly,ke2021musiq}, triggers based on a local patch or predefined global patterns may not persist {after} the cropping or data augmentations during training.
To address the diminishing effect of triggers during training/attacking, we propose adding the trigger with a patch-based pattern across the global area. By introducing the trigger in the \textbf{DCT} space, which is inherently block-based, a patch-based pattern is effectively created across the entire image, making it well-suited for our attacks.

\noindent \textbf{Trigger Injection.} 
The trigger injection model $T\left(\boldsymbol{x}, {\alpha}\cdot\boldsymbol{t}\right)$ takes a image $\bm{x}$, the trigger $\boldsymbol{t}$, and the scaling coefficient $\alpha$ to {generate} a poisoned image $\bm{x_p}$. 
Given the prevalence of DCT in coding techniques~\cite{wallace1992jpeg} and its ability to perform patch-based transforms on images of varying sizes, we propose a frequency-based trigger injection.
Given $\bm{x}$, we split it into non-overlapping patches $\bm{x}_{patch}$. Following a 2d-DCT transform, we have the $\bm{x}_{dct}$. By adding the scaled trigger ${\alpha}\!\cdot\!\boldsymbol{t}$ to all patches of $\bm{x}_{dct}$, we have the triggered $\bm{x}_{dct}^{p}$.
The final result $T\left(\boldsymbol{x}, {\alpha}\!\cdot\!\boldsymbol{t}\right)$ is obtained by applying an inverse 2d-DCT. 
The overall process is given by:
$
\forall\alpha\!\in\![-1,1]\!: \boldsymbol{x}_p \!=\! T\left(\boldsymbol{x}, {\alpha}\!\cdot\!\boldsymbol{t}\right) \!=\! \text{IDCT}(\text{DCT}(\boldsymbol{x})\!+\!{\alpha}\!\cdot\!\boldsymbol{t}),
$
where DCT and inverse-DCT (IDCT) {are} on all patches.
To maintain imperceptibility, we add in {middle} frequencies,
as adding in low frequencies induces significant perturbations in the spatial domain,
while high frequencies are less robust. 
We use a patch size of $16\!\times\!16$, and perturb the mid-64 frequencies.


\begin{algorithm}[t]
\caption{Searching UAP in the DCT domain (UAP-DCT)}
\begin{algorithmic}[1]
\REQUIRE Subset $\mathcal{D}_s = \{(\boldsymbol{x}_i, y_i)\}_{i=1}^{N_s}$, surrogate model $f_{\boldsymbol{\theta}_s}$, perturbation bound $\epsilon$, epochs $e_1$, $e_2$
\ENSURE Optimized trigger $\boldsymbol{t}$
\STATE Train $f_{\boldsymbol{\theta}_s}$ on $\mathcal{D}_s$ using loss $\mathcal{L}_1$ and Adam for $e_1$ epochs
\FOR{$i = 1$ \TO $e_2$}
    \FOR{each $(\boldsymbol{x}, y) \in \mathcal{D}_s$}
        \STATE $\mathcal{L}_{UAP} = - \underbrace{\lVert f_{\boldsymbol{\theta}_s}(T(\boldsymbol{x}, \boldsymbol{t})) - f_{\boldsymbol{\theta}_s}(\boldsymbol{x}) \rVert_1}_{\text{effectiveness of UAP-DCT}}$\par\hskip\algorithmicindent $
         \quad\quad\quad + \lambda \cdot \underbrace{\max(\mathcal{L}_{mse}(\boldsymbol{x}, T(\boldsymbol{x}, \boldsymbol{t})), \epsilon^2)}_{\text{invisibility of UAP-DCT}}
        $
        \STATE Update $\boldsymbol{t}$ by minimizing $\mathcal{L}_{UAP}$ using the Adam
    \ENDFOR
\ENDFOR
\end{algorithmic}
\label{alg:UAP-DCT}
\end{algorithm}

\noindent \textbf{Exploiting DCT Space Vulnerability.} Indeed, research has revealed that state-of-the-art DNN classifiers are vulnerable to a phenomenon known as Universal Adversarial Perturbations (UAP), which refers to a universal and invisible perturbations that can cause high-probability misclassifications.
Inspired by the efficacy of universally applicable perturbations in misleading well-trained models, we leverage UAP in the DCT space, denoted as UAP-DCT, as the trigger $\boldsymbol{t}$. 
By incorporating these UAP-DCT triggers into a poisoned dataset, the attacker can amplify the susceptibility of a trained IQA model to such triggers, thereby enhancing {attack} performance and facilitating more effective manipulation of the attacked outputs. 
Algorithm~\ref{alg:UAP-DCT} outlines the optimization process for the trigger $\boldsymbol{t}$, involving two loss terms. 
The first loss aims to generate effective UAP-DCT, while the second loss regulates the magnitude of UAP-DCT to ensure the induced perturbations remain invisible in the spatial domain, thereby enhancing the imperceptibility of the poisoned images.

\subsection{Backdoor Attacks with Poison Label}
The poison-label backdoor attacks against NR-IQA (\textbf{P-BAIQA}) are introduced to concretely validate the potency of our designed backdoor triggers and attacking goals. 
Specifically, similar to the existing poison-label backdoor attacks, our P-BAIQA first randomly select a subset $\mathcal{D}_s$ with poisoning ratio $r$ from the benign dataset $\mathcal{D}$ to make its modified version $\mathcal{D}_{p} = \{ (T(\boldsymbol{x}, {\alpha} \cdot \boldsymbol{t}), y + \alpha \cdot {\Delta{y}}_{t})|\alpha \sim \mathcal{P}_{\alpha}, (\bm{x},y) \in \mathcal{D}_s\}$, where $\mathcal{P}_{\alpha}$ denotes sampling distribution of $\alpha$. 
Prior to generating the poisoned dataset, we optimize the trigger $\boldsymbol{t}$ based on $\mathcal{D}_s$ using Algorithm~\ref{alg:UAP-DCT}.
Given the sampling distribution $\mathcal{P}_{\alpha}$, we also explore several cases: 1) sampling from {{$\{\pm1,\pm\frac{3}{4},\pm\frac{1}{2},\pm\frac{1}{4}\}$}} with probabilities {{$\{0.4,0.3,0.2,0.1\}$}}; 2) sampling from {{$\{\pm1\}$}} with equal probability; 3) Uniform distribution (i.e., {{$\mathrm{U}(-1,1)$}}).
The modified subset $\mathcal{D}_{p}$ with the remaining clean samples $\mathcal{D}_{c} = \mathcal{D}\backslash \mathcal{D}_s$ will then be released to train the model $f_{\boldsymbol{\theta}}(\cdot)$ as indicated in Eq.~\ref{eq:backdoor_attack}.

During the inference, for any test sample $(\bm{x},y)$, the adversary can trigger the backdoor using the poisoned $T\left(\boldsymbol{x}, {\alpha}\cdot\boldsymbol{t}\right)$. The adversary is free to choose the $\alpha$, thus manipulating the output to the desired target $f_{\boldsymbol{\theta}^{*}}( \boldsymbol{x}) + \alpha \cdot {\Delta{y}}_{t}$ as intended.

\subsection{Backdoor Attacks with Clean Label}\label{sec:clean_label}
As we will demonstrate in Section 4, the heuristic P-BAIQA can achieve promising results. However, it \textbf{lacks stealthiness}, as it retains poisoned labels. Dataset users may detect the poisoned data through label inspection. In this section, we explore the design of clean-label backdoor attacks (\textbf{C-BAIQA}), focusing on the sampling strategy of $\alpha$ and the modification of $\boldsymbol{x}$.
We first present the necessary assumption and theorem.
%
\subsubsection{Assumption 1}\label{assumption1}
\textit{For a backdoored model $f_{\boldsymbol{\theta}}$, we assume that for any input and the true label $(\boldsymbol{x},y) \!\sim\! \mathcal{C}$, the output $\tilde{y}$ follows a Gaussian distribution with an variance $\sigma^2$:}
\begin{equation}\small
\begin{split}
&\mathbb{P}(\tilde{y}|\bm{x},\alpha) = \mathcal{N}(\tilde{y};y + \alpha{\Delta{y}}_{t},\sigma^2), ~\mathbb{P}(\tilde{y}|\bm{x})=\mathcal{N}(\tilde{y};y,\sigma^2),\\
&\mathbb{P}(\tilde{y}) = \mathcal{N}(\tilde{y};\mu_{\tilde{y}},\sigma^2), ~\mathbb{P}(\tilde{y}|\alpha)=\mathcal{N}(\tilde{y};\mu_{\tilde{y}|\alpha},\sigma^2),
\end{split}
\label{poisoned_distribution}
\end{equation}
\textit{where $\mathcal{C}$ is the clean test data, and $\mu_{\tilde{y}}$, $\mu_{\tilde{y}|\alpha}$ are given by}
\begin{equation}\small
\begin{split}
    \mu_{\tilde{y}|\alpha} = \mathbb{E}[\tilde{y}|\alpha] = \mathbb{E}_{\bm{x}}[\mathbb{E}[\tilde{y}|\bm{x},\alpha]]=\mathbb{E}_{\bm{x}}[\mathbb{E}[\tilde{y}|\bm{x}]]+\alpha \cdot {\Delta{y}}_{t}\\=\mathbb{E}[\tilde{y}]+\alpha \cdot {\Delta{y}}_{t}=\mu_{\tilde{y}}+\alpha \cdot {\Delta{y}}_{t}.
\end{split}
\end{equation}
\textit{For clean-label attacks, where labels in $\mathcal{D}_{p}$ remain unchanged, we assume that $f{\boldsymbol_{\bm{\theta}}}$ is capable of learning the label priors during the training process. Thus, we have:}
\begin{equation}\small
\begin{split}
\mu_{\tilde{y}}=\mathop\mathbb{E}\limits_{y\sim\mathcal{D}_{t}}[y]=\mu_y, ~\mu_{\tilde{y}|\alpha}=\mu_{\tilde{y}}+\alpha {\Delta{y}}_{t}=\mu_y+\alpha{\Delta{y}}_{t}.
\end{split}
\label{poisoned_label}
\end{equation}

Assumption 1 aligns with the fundamental characteristics of backdoor attacks at test time. 
Specifically, $\mathbb{P}(\tilde{y}|\bm{x},\!\alpha)$ is the output distribution when triggered data is input, while $\mathbb{P}(\tilde{y}|\bm{x})$ is the output distribution for clean data. 
Additionally, $\mathbb{P}(\tilde{y})$ serves as the prior for this task, and $\mathbb{P}(\tilde{y}|\bm{x},\!\alpha)$ reflects the conditional distribution when considering only the trigger. 
The means of these distributions are consistent with backdoor attack behavior. Furthermore, it is common to assume that the output variations are identical, denoted as $\sigma^2$. 

For Eq.~\ref{poisoned_label}, in a simplified context, $\mu_{\tilde{y}}$ serves as the prior of the trained model and is generally understood as the expected value of the labels across the entire training dataset. Conversely, $\mu_{\tilde{y}|\alpha}$ is the prior of the trained model in the presence of the trigger, typically corresponding to the expected value of the labels within the poisoned subset of the training dataset.
Therefore, based on this, we can develop the sampling strategy for $\alpha$ as outlined in the following remark.

\subsubsection{Remark 1 (Sampling strategy of $\alpha$ in $\mathcal{D}_p$)}\label{remark1}
\textit{To train a model $f_{\boldsymbol{\theta}}$ satisfying the Eq.~\ref{poisoned_label}, the poisoned training set $\mathcal{D}_p$ should adhere to those properties as well. Thus, considering the distribution of the label $y$, one approach to sample $\alpha$ satisfying Eq.~\ref{poisoned_label} is to select $\alpha = (y - \mu_y)/{{\Delta{y}}_{t}}$ for $\mathcal{D}_p$.}

Remark 1 outlines the sampling strategy for $\alpha$ based on Assumption 1. Additionally, for Eq.~\ref{poisoned_distribution}, it presents the mathematical relationship $\mathbb{P}(\tilde{y}|\bm{x},\alpha)\mathbb{P}(\tilde{y})=\mathbb{P}(\tilde{y}|\bm{x})\mathbb{P}(\tilde{y}|\alpha)$ for any test data. Since $f_{\bm{\theta}}$ learns this property from the training set, it is crucial that the training set adheres to this equation. To increase the likelihood of this equation and ensure consistency between training on $\mathcal{D}_p$ and evaluation during the attack phase, we consider whether modifying $\bm{x}$ to $\bm{x^{\prime}}$ is necessary. This leads to the development of the following theorem.

\begin{algorithm}[t]
\caption{C-BAIQA}
\begin{algorithmic}[1]
\REQUIRE Benign dataset $\mathcal{D} = \{(\boldsymbol{x}_i, y_i)\}_{i=1}^N$, surrogate model $f_{\boldsymbol{\theta}_s}$, poisoning ratio $r$, hyperparameters for UAP-DCT ($\epsilon$, $e_1$, $e_2$), hyperparameters for TAEs (target $\mu_y$, bound $\epsilon_t$, step size $\alpha_t$, iterations $I_t$)
\ENSURE Optimized trigger $\boldsymbol{t}$, Modified dataset $\mathcal{D}_t$
\STATE $\mathcal{D}_s \leftarrow$ Randomly sample a subset from $\mathcal{D}$ with ratio $r$
\STATE \textcolor{teal}{\# Train surrogate model and search trigger}
\STATE $f_{\boldsymbol{\theta}_s}, \boldsymbol{t} \leftarrow$ Apply Algorithm~\ref{alg:UAP-DCT} with $(\mathcal{D}_s, f_{\boldsymbol{\theta}_s}, \epsilon, e_1, e_2)$
\STATE \textcolor{teal}{\# Generate poisoned subset $\mathcal{D}_p$}
\FOR{each $(\boldsymbol{x}_i, y_i) \in \mathcal{D}_s$}
    \STATE $\alpha_i \leftarrow (y_i - \mu_y) / \Delta{y}_t$
    \STATE $\boldsymbol{x}_i^{\prime} \leftarrow$ \textbf{Targeted-PGD}(model=$f_{\boldsymbol{\theta}_s}$, input=$\boldsymbol{x}_i$, \par\hskip\algorithmicindent \quad\quad\quad\quad~~ target=$\mu_y$, bound=$\epsilon_t$, step=$\alpha_t$, iter=$I_t$)
\ENDFOR
\STATE $\mathcal{D}_p \leftarrow \{(T(\boldsymbol{x}_i^{\prime}, g(\alpha_i) \cdot \boldsymbol{t}), y_i)\}_{i=1}^{N_s}$
\STATE $\mathcal{D}_c \leftarrow \mathcal{D} \setminus \mathcal{D}_s$, $\mathcal{D}_t \leftarrow \mathcal{D}_c \cup \mathcal{D}_p$
\RETURN $\boldsymbol{t}$, $\mathcal{D}_t$
\end{algorithmic}
\label{alg:clean_label}
\end{algorithm}

\subsubsection{Theorem 1}\label{theorem1}
\textit{For any data point $(\boldsymbol{x^{\prime}},y) \in \mathcal{D}_p$, consistency between training and testing requires that it satisfies the distributions in Eq.~\ref{poisoned_distribution}. This condition can be expressed as}
\begin{equation}\small
\begin{split}
&\quad\quad\quad\mathbb{P}(\tilde{y}|\bm{x^{\prime}},\alpha)\mathbb{P}(\tilde{y})=\mathbb{P}(\tilde{y}|\bm{x^{\prime}})\mathbb{P}(\tilde{y}|\alpha),
\end{split}
\end{equation}
\textit{which is identical to}
\begin{equation}\small
\begin{split}
\mathbb{P}(\tilde{y}|\bm{x^{\prime}},\alpha)= \frac{\mathbb{P}(\tilde{y}|\alpha)}{\mathbb{P}(\tilde{y})}\mathbb{P}(\tilde{y}|\bm{x^{\prime}})=\frac{\mathbb{P}(\alpha|\tilde{y})}{\mathbb{P}(\alpha)}\mathbb{P}(\tilde{y}|\bm{x^{\prime}}).
\end{split}
\end{equation}
\textit{Considering Bayes' theorem, we also derive}
\begin{equation}\small
\begin{split}
\!\!\mathbb{P}(\tilde{y}|\bm{x^{\prime}},\alpha)\!=\! \frac{\mathbb{P}(\alpha|\bm{x^{\prime}},\tilde{y})\mathbb{P}(\bm{x^{\prime}}|\tilde{y})\mathbb{P}(\tilde{y})}{\mathbb{P}(\alpha|\bm{x^{\prime}})\mathbb{P}(\bm{x^{\prime}})}\!=\!\frac{\mathbb{P}(\alpha|\bm{x^{\prime}},\tilde{y})}{\mathbb{P}(\alpha|\bm{x^{\prime}})}\mathbb{P}(\tilde{y}|\bm{x^{\prime}}).
\end{split}   
\end{equation}
\textit{By comparing the two expressions for $\mathbb{P}(\tilde{y}|\bm{x^{\prime}},\alpha)$, we have}
\begin{equation}\small
\begin{split}
\!\!\frac{\mathbb{P}(\alpha|\tilde{y})}{\mathbb{P}(\alpha)} = \frac{\mathbb{P}(\alpha|\bm{x^{\prime}},\tilde{y})}{\mathbb{P}(\alpha|\bm{x^{\prime}})}\overset{Bayes}{\iff}\frac{\mathbb{P}(\alpha|\tilde{y})}{\mathbb{P}(\alpha)} =\frac{\mathbb{P}(\alpha|\tilde{y})\mathbb{P}(\bm{x^{\prime}}|\alpha,\tilde{y})}{\mathbb{P}(\alpha|\bm{x^{\prime}})\mathbb{P}(\bm{x^{\prime}}|\tilde{y})}.
\end{split}
\label{property_converted_x}
\end{equation}

\vspace{2mm}
Theorem 1 highlights the necessity of modifying $\bm{x}$ to $\bm{x^{\prime}}$ within $\mathcal{D}_p$. Consequently, the generation of $\mathcal{D}_p$ can be divided into two steps. First, after selecting the subset $\mathcal{D}_s$, $\boldsymbol{x}$ is transformed into $\boldsymbol{x^{\prime}}$. Second, the sampling strategy outlined in Remark 1 is applied to convert $\boldsymbol{x^{\prime}}$ into $T(\boldsymbol{x^{\prime}}, {\alpha} \cdot \boldsymbol{t})$.

\subsubsection{Remark 2  (Strategy of converting $\boldsymbol{x}$ into $\boldsymbol{x^{\prime}}$)}\label{remark2}
\textit{To convert $\boldsymbol{x}$ into $\boldsymbol{x^{\prime}}$ satisfying Eq.~\ref{property_converted_x}, one approach is to ensure $\boldsymbol{x^{\prime}}$ is independent of both $\alpha$ and $\tilde{y}$. Since sampling of $\alpha$ is based on $y$, making $\boldsymbol{x^{\prime}}$ independent of $\tilde{y}$ is sufficient. In this case, we have $\mathbb{P}(\tilde{y}|\bm{x^{\prime}})=\mathbb{P}(\tilde{y})=\mathcal{N}(\tilde{y};\mu_{\tilde{y}},\sigma^2)$, following the distribution in Eq.~\ref{poisoned_distribution}.
Thus, by setting targeted adversarial examples (TAEs) of $\boldsymbol{x}$ (with $\mu_y$ as the target score) as the modified $\bm{x^{\prime}}$, the above equation is satisfied. 
Specifically, since trigger injection occurs in the DCT space, we choose the spatial domain as the perturbation space for generating $\boldsymbol{x^{\prime}}$.}

The overall framework for C-BAIQA is in Algorithm~\ref{alg:clean_label}. 
After randomly {selecting} a subset $\mathcal{D}_s$ from $\mathcal{D}$, we first use Algorithm~\ref{alg:UAP-DCT} to train $f_{\boldsymbol{\theta}_s}$ and search $\bm{t}$.
Following Remark 1 and Remark 2, we then transform $\mathcal{D}_s$ into $\mathcal{D}_p$, and mix it with the remaining $\mathcal{D}_c$ to construct the final training set $\mathcal{D}_t$.

\begin{table*}[t]\small
\centering
\scalebox{1.0}{
\setlength{\tabcolsep}{0.9pt}
\begin{tabular}{c c||*3{c}|*2{c}|*3{c}|*2{c}||*3{c}|*2{c}|*3{c}|*2{c}}
\hline
\multicolumn{2}{c||}{Attack Settings} &
\multicolumn{10}{c||}{Poison-label attacks (P-BAIQA)} &
\multicolumn{10}{c}{Clean-label attacks (C-BAIQA)} \\
\hline
\multicolumn{2}{c||}{Dataset} &
\multicolumn{5}{c|}{LIVEC} &
\multicolumn{5}{c||}{KonIQ-10k}&
\multicolumn{5}{c|}{LIVEC} &
\multicolumn{5}{c}{KonIQ-10k}\\
\hline
\multirow{2}{*}{Model}& \multirow{2}{*}{Attacks}&\multicolumn{3}{c|}{Benign}&\multicolumn{2}{c|}{Attack}&\multicolumn{3}{c|}{Benign}&\multicolumn{2}{c||}{Attack}&\multicolumn{3}{c|}{Benign}&\multicolumn{2}{c|}{Attack}&\multicolumn{3}{c|}{Benign}&\multicolumn{2}{c}{Attack}
\\
&& \textcircled{a} & \textcircled{b} & \textcircled{c} & \textcircled{{A}} & \textcircled{{B}}& \textcircled{a} & \textcircled{b} & \textcircled{c} & \textcircled{{A}} & \textcircled{{B}}& \textcircled{a} & \textcircled{b} & \textcircled{c} & \textcircled{{A}} & \textcircled{{B}}& \textcircled{a} & \textcircled{b} & \textcircled{c} & \textcircled{{A}} & \textcircled{{B}}\\
\cline{1-22}
\multirow{5}{*}{\rotatebox{90}{HyperIQA}} & w/o & 0.911 & 0.897 & 9.47 & - & - & 0.905 & 0.886 & 7.00 & - & - & 0.911 & 0.897 & 9.47 & - & - & 0.905 & 0.886 & 7.00 & - & -\\
& Blended & 0.887 & 0.859 & 10.27 & 18.88 & 0.11 & 0.894 & 0.876 & 7.31 & 18.09 & 0.14 & 0.903 & 0.883 & 9.91 & 22.47 & -0.02 & 0.899 & 0.885 & 7.26 & 22.25 & -0.00\\
& WaNet & 0.877 & 0.855 & 10.59 & 22.03 & -0.00 & 0.893 & 0.878 & 7.09 & 22.01 & -0.00 & 0.877 & 0.855 & 10.59 & 22.03 & -0.00 & 0.909 & 0.895 & 6.66 & 22.00 & 0.00\\
& FTrojan & 0.912 & 0.878 & 9.64 & 10.18 & 0.51 & 0.897 & 0.879 & 7.10 & 7.77 & 0.63 & 0.905 & 0.883 & 9.40 & 18.92 & 0.13 & 0.905 & 0.894 & 7.00 & 16.31 & 0.30\\
& Ours & 0.903 & 0.872 & 9.52 & \textbf{8.72} & \textbf{0.63} & 0.903 & 0.887 & 6.93 & \textbf{6.42} & \textbf{0.65} & 0.903 & 0.892 & 9.91 & \textbf{10.65} & \textbf{0.60} & 0.900 & 0.885 & 7.00 & \textbf{15.19} & \textbf{0.29}\\
\hline
\multirow{5}{*}{\rotatebox{90}{DBCNN}} & w/o & 0.878 & 0.869 & 10.48 & - & - & 0.905 & 0.889 & 6.74 & - & - & 0.878 & 0.869 & 10.48 & - & - & 0.905 & 0.889 & 6.74 & - & -\\
& Blended & 0.855 & 0.817 & 10.96 & 18.44 & 0.13 & 0.887 & 0.865 & 7.34 & 14.00 & 0.29 & 0.773 & 0.717 & 17.01 & 22.18 & -0.01 & 0.888 & 0.881 & 7.47 & 22.27 & -0.01\\
& WaNet & 0.828 & 0.798 & 11.71 & 22.04 & -0.00 & 0.867 & 0.839 & 7.78 & 21.96 & 0.00 & 0.828 & 0.798 & 11.71 & 22.04 & -0.00 & 0.891 & 0.885 & 7.38 & 22.00 & -0.00\\
& FTrojan & 0.866 & 0.848 & 10.90 & 10.50 & 0.50 & 0.898 & 0.878 & 7.11 & 5.60 & 0.67 & 0.771 & 0.723 & 17.15 & 21.89 & 0.00 & 0.890 & 0.882 & 7.44 & 21.12 & 0.04\\
& Ours & 0.887 & 0.860 & 9.98 & \textbf{9.45} & \textbf{0.56} & 0.902 & 0.883 & 7.03 & \textbf{5.21} & \textbf{0.69} & 0.890 & 0.866 & 10.09 & \textbf{12.37} & \textbf{0.45} & 0.903 & 0.887 & 6.95 & \textbf{13.97} & \textbf{0.37}\\
\hline
\multirow{5}{*}{\rotatebox{90}{TReS}} & w/o & 0.909 & 0.894 & 15.22 & - & - & 0.909 & 0.886 & 19.20 & - & - & 0.909 & 0.894 & 15.22 & - & - & 0.909 & 0.886 & 19.20 & - & -\\
& Blended & 0.878 & 0.845 & 17.60 & 20.15 & 0.10 & 0.871 & 0.860 & 15.08 & 19.49 & 0.10 & 0.897 & 0.872 & 17.61 & 23.08 & -0.04 & 0.897 & 0.880 & 18.22 & 22.86 & -0.03\\
& WaNet & 0.881 & 0.869 & 16.80 & 22.00 & 0.00 & 0.839 & 0.836 & 24.14 & 22.08 & -0.00 & 0.881 & 0.869 & 16.80 & 22.00 & 0.00 & 0.904 & 0.887 & 18.94 & 21.99 & 0.00\\
& FTrojan & 0.866 & 0.843 & 18.11 & 10.33 & 0.66 & 0.891 & 0.877 & 20.76 & 9.81 & 0.81 & 0.895 & 0.882 & 17.94 & 19.60 & 0.10 & 0.886 & 0.872 & 17.24 & 16.96 & 0.22\\
& Ours & 0.877 & 0.855 & 17.31 & \textbf{10.22} & \textbf{0.73} & 0.892 & 0.875 & 22.24 & \textbf{9.01} & \textbf{0.93} & 0.895 & 0.871 & 16.15 & \textbf{13.75} & \textbf{0.42} & 0.887 & 0.862 & 16.99 & \textbf{14.60} & \textbf{0.35}\\
\hline
\end{tabular}}
\caption{Comparison of P-BAIQA and C-BAIQA with baseline attack methods with the poisoning ratio $r=20\%$.
\textcircled{a}, \textcircled{b} and \textcircled{c} denote the PLCC, SROCC and RMSE, respectively.
\textcircled{{A}} and \textcircled{{B}} denote the mMAE $\downarrow$ and mMRA $\uparrow$, respectively.
}
\label{table:poison_label_comparison}
\end{table*}

\begin{table}[t]\small
\centering
\setlength{\tabcolsep}{8.0pt}
\begin{tabular}{c|c|c|c|c}
\hline
Attacks~$\rightarrow$ & Blended & WaNet & FTrojan & Ours \\
\hline
LIVEC & 27.71 & 26.59 & 29.80 & \textbf{30.06} \\
KonIQ-10k & 33.00 & 34.81 & 35.91 & \textbf{36.32} \\
\hline
\end{tabular}
\caption{Imperceptibility of the poisoned data ($\text{PSNR}_{1} \uparrow$).}
\label{tab:psnr}
\end{table}

\section{Experiments}
\subsection{Experimental Setup}\label{sec:set_up}
\noindent\textbf{Datasets and Models.} 
We choose the LIVEC dataset~\cite{ghadiyaram2015massive} consisting of 1162 images with sizes $500\times500$ and KonIQ-10k dataset~\cite{hosu2020koniq} consisting of 10073 images with sizes $512\times384$. For LIVEC, consistent with prior research~\cite{liu2024defense}, we randomly select 80\% of the images as the training set, and the remaining 20\% as the testing set.
For KonIQ-10k, we adopt the official split with 70\% of the images as the training set. 
We include two CNN-based NR-IQA models: HyperIQA~\cite{su2020blindly}, and DBCNN~\cite{zhang2020blind}, and one Transformer-based TReS~\cite{golestaneh2021no}. 
%

\noindent\textbf{Baseline Attack Methods.} We compare our attacks with existing poisoning-based BA.
We utilize Blended~\cite{chen2017targeted}, WaNet~\cite{nguyen2021wanet}, and FTrojan~\cite{wang2021backdoor}, which are representative non-patch-based invisible attacks.
For all baselines, we employ the coefficient $\alpha$ to adjust the intensity and strength of the triggers, following a similar approach to our attacks.
For FTrojan, we add the triggers into the same mid-frequency channels.
We apply {the} same configurations to all methods, except the trigger. 

\noindent\textbf{Evaluation Metrics.}
For the evaluation on benign data, we follow previous works and consider three metrics, \textit{i.e.,} RMSE, PLCC, and SROCC.
To evaluate the attack effectiveness, 
we include the mean absolute error (MAE), and the mean ratio of amplification (MRA) across the test set regarding each $\alpha \!\!\in\!\! [-1,1]$. 
While it is impossible to consider all values of $\alpha$, we estimate it using finite $\alpha$ from a selected set $A$. Therefore, we utilize the mean of {those} metrics for an overall evaluation:
\begin{equation}\small
\begin{split}
&\text{MAE}(\alpha) = \mathop\mathbb{E}\limits_{(\boldsymbol{x},y) \sim \mathcal{C}}\Big[\big\lvert \Delta_{{a}}(\boldsymbol{x},\alpha) -\Delta_{{t}}(\alpha)\big\lvert\Big],\\
&\text{MRA}(\alpha) =\mathop\mathbb{E}\limits_{(\boldsymbol{x},y) \sim \mathcal{C}}\Big[\frac{\Delta_{{a}}(\boldsymbol{x},\alpha)}{ \Delta_{{t}}(\alpha)}\Big],\\
&\text{mMAE} = \frac{1}{n(A)}\sum_{\alpha \in A}\text{MAE}(\alpha),
\text{mMRA} = \frac{1}{n(A)}\sum_{\alpha \in A}\text{MRA}(\alpha), \\
&\text{with}~\Delta_{{a}}(\boldsymbol{x},\alpha) = f_{\boldsymbol{\theta}^{*}}( T(\boldsymbol{x}, {\alpha} \cdot \boldsymbol{t})) - f_{\boldsymbol{\theta}^{*}}( \boldsymbol{x}), \Delta_{{t}}(\alpha) = \alpha \cdot {\Delta{y}}_{t}.\nonumber
\end{split}
\end{equation}
Experimentally, we choose {\small{$A \!\!=\!\! \{\pm0.1, \pm0.2, ..., \pm1.0\}$}}.
For the {imperceptibility of} poisoned images, we provide the PSNR between $\boldsymbol{x_p}$ and $\boldsymbol{x}$, denoted as $\text{PSNR}_{1}$, when we set $\alpha\!\!=\!\!1$.

\subsection{Performance of P-BAIQA}
\noindent\textbf{Settings.}
As the MOS range is $[0,100]$, we set the maximum deviation $\Delta{y_t}=40$, ensuring this deviation is sufficient to yield potent attacks.
To regulate the imperceptibility of poisoned images, in Algorithm~\ref{alg:UAP-DCT}, we set {{$\lambda=10^8$}}, {{$e_1=24$}}, {{$e_2=50$}}, and $\epsilon=\frac{8}{255}$ for LIVEC. For KonIQ-10k, where a larger number of images are eligible for poisoning, we adjust $\epsilon$ to a lower value $\frac{4}{255}$.
We set the poisoning ratio $r=20\%$ for all attacks and datasets, and use the HyperIQA as the surrogate model. 
For the sampling {strategy} of $\alpha$ to construct $\mathcal{D}_p$, we consider sampling from {{$\{\pm1,\pm\frac{3}{4},\pm\frac{1}{2},\pm\frac{1}{4}\}$}} with probabilities {{$\{0.4,0.3,0.2,0.1\}$}}, and provide results of other strategies in the ablation study.

\noindent\textbf{Results.} 
As depicted in Table~\ref{table:poison_label_comparison} and Table~\ref{tab:psnr}, our approach consistently surpasses the baseline attacks on both datasets, demonstrating both a more potent attack effectiveness and an enhanced level of invisibility for the trigger.
Blended and WaNet encounter significant difficulties when attempting to attack NR-IQA models.
This is likely due to two key factors: 1) Blended adds a predefined trigger across the entire image area. However, during the training and testing of NR-IQA models, a random cropping operation is typically applied, which disrupts the direct correlation between the global spatial trigger and the variations in the model's output. 
2) WaNet employs a wrapping function to inject the trigger, but this method may not be effectively recognized or learned by NR-IQA models, as their output space is continuous and the target score to be manipulated can be any value.
In contrast to FTrojan, which relies on manually defined triggers in the DCT space, our approach leverages UAP, capitalizing on the vulnerabilities in {the} DCT space to amplify the effectiveness of the attack.
Moreover, the benign metrics also show that our attack have low performance impact on the clean data. 

\begin{figure}[t]
\centering
\includegraphics[width=0.95\linewidth]{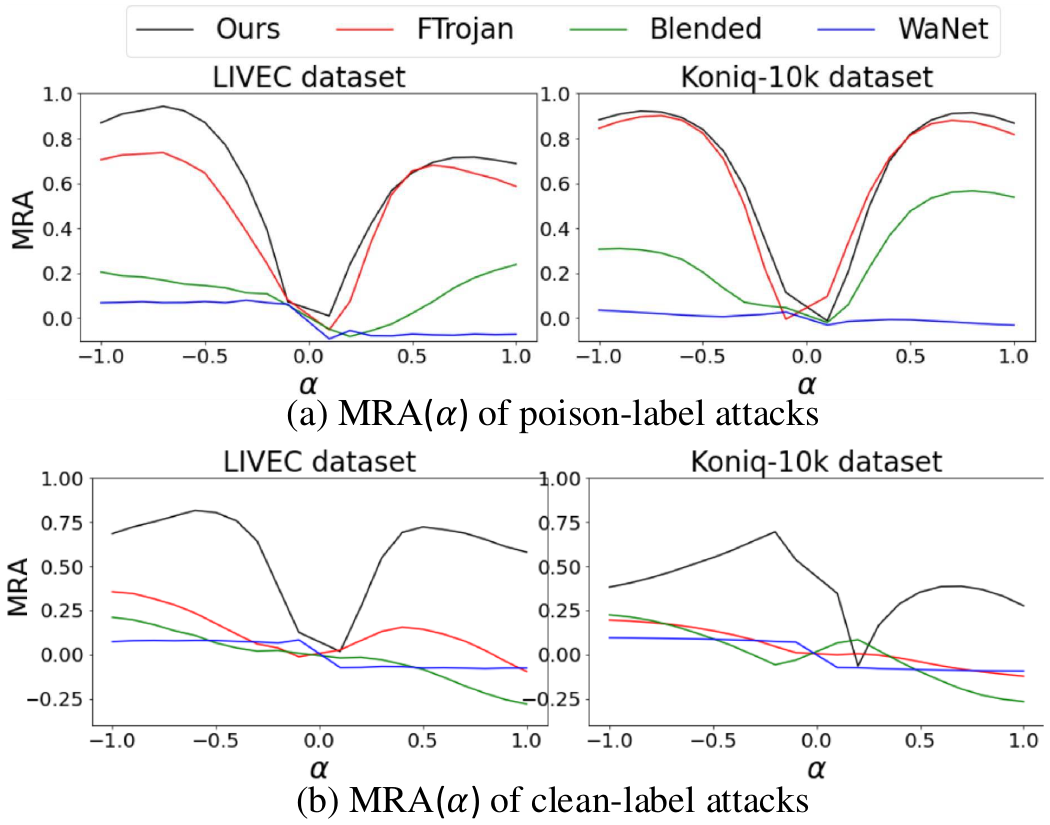}
\caption{
$\text{MRA}(\alpha)$ with HyperIQA as victim models.
}
\label{mra}
\end{figure}

\begin{figure*}[t]
\centering
\includegraphics[width=0.98\linewidth]{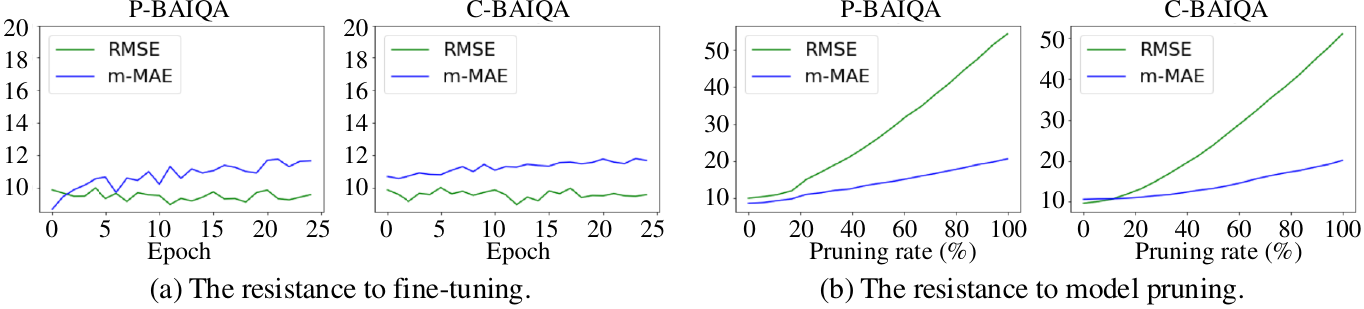}
\caption{
Resistance to fine-tuning and pruning (HyperIQA as models and LIVEC as the dataset).
}
\label{resistance}
\end{figure*}

\begin{table*}[t]\small
\centering
\scalebox{1.0}{
\setlength{\tabcolsep}{3pt}
\begin{tabular}{c c||*2{c}|*2{c}|*2{c}||*2{c}|*2{c}|*2{c}}
\hline
\multirow{3}{*}{\shortstack{Attack\\Type}}&{Dataset $\rightarrow$} &
\multicolumn{6}{c||}{LIVEC} &
\multicolumn{6}{c}{KonIQ-10k}\\
\cline{2-14}
&{Model$\rightarrow$}&\multicolumn{2}{c|}{HyperIQA}&\multicolumn{2}{c|}{DBCNN}&\multicolumn{2}{c||}{TReS}&\multicolumn{2}{c|}{HyperIQA}&\multicolumn{2}{c|}{DBCNN}&\multicolumn{2}{c}{TReS}
\\
&{Metrics$\rightarrow$}& mMAE & mMRA&mMAE & mMRA&mMAE & mMRA&mMAE& mMRA&mMAE& mMRA&mMAE& mMRA\\

\cline{1-14}
\multirow{3}{*}{{P-BAIQA}} & \ding{172} & \textbf{7.825} & \textbf{0.7660} & \textbf{9.093} & \textbf{0.7900} & 11.241 & 0.8244 & 7.548 & \textbf{0.8540}  & 6.741 & \textbf{0.8078} & 13.193 & \textbf{1.0916}\\
& \ding{173} & 9.485&0.5419 & 9.608&0.5756& 11.831&0.6942&7.029&0.6413&5.510&0.6813&{9.543}&0.8308 \\
& Ours&  8.719 & 0.6344& 9.449 & 0.5611 & \textbf{10.223} & \textbf{0.7274} &\textbf{6.420} &{0.6487}& \textbf{5.208}& {0.6917} & \textbf{9.012} &0.9291\\
\hline
\multirow{3}{*}{{C-BAIQA}}& \ding{174} & 12.689 & 0.4117 & 12.610 & 0.4195 & 14.597 & 0.3490 & \textbf{15.166} & \textbf{0.3002}  & 14.277 & 0.3463 & 17.837 & 0.1714\\
& \ding{175} & 20.777& 0.0638& 19.320 & 0.1257 & 19.089 & 0.1271&19.023&0.1924&16.388&0.3242&21.499&0.0249 \\
&Ours  &  \textbf{10.653} &\textbf{0.5983} & \textbf{12.372} &\textbf{0.4525} & \textbf{13.746} &\textbf{0.4183} &15.185& 0.2859& \textbf{13.970}& \textbf{0.3682}& \textbf{14.597} & \textbf{0.3490}\\
\hline
\end{tabular}
}
\caption{Ablation study of our attacks. 
\ding{172} and \ding{173} denote our P-BAIQA with $\alpha$ sampling from $\{\pm1\}$ and $\mathrm{U}(-1,1)$, respectively.
\ding{174} denotes our C-BAIQA without converting $\boldsymbol{x}$ into $\boldsymbol{x^{\prime}}$.
\ding{175} denotes our C-BAIQA by setting $\mu_y\!\!=\!\!65$ for the sampling strategy of $\alpha$.}
\label{table:ablation}
\end{table*}

In addition, we provide the $\text{MRA}(\alpha)$ in Fig.~\ref{mra}(a). We can see that our method is capable of achieving a significant deviation when $\alpha$ has an absolute value greater than 0.5. However, when $\alpha$ {falls} within [-0.5, 0.5], the manipulation becomes less precise, as the model finds it more challenging to recognize the trigger.
However, we believe this limitation is tolerable given that attackers generally aim for significant shifts in outputs rather than minor alterations.
%

\subsection{Performance of C-BAIQA}
\noindent\textbf{Settings.} The configurations for $\lambda$, $e_1$, $e_2$, $\epsilon$, $\Delta{y_t}$, $r$, and the surrogate model in Algorithm~\ref{alg:UAP-DCT} match those in P-BAIQA.
As the MOS range is $[0,100]$, we set $\mu_y=50$ in Algorithm~\ref{alg:clean_label}.
For the TAEs, we adopt PGD attacks~\cite{DBLP:conf/iclr/MadryMSTV18}, with $\epsilon_t$ in terms of $\ell_{\infty}$-norm (\textit{i.e.,} $\epsilon_t=2/255$ for LIVEC and $\epsilon_t=1/255$ for  KonIQ-10k), $\alpha_t=1/255$, and $I_t=20$.

\noindent\textbf{Results.}
From Table~\ref{table:poison_label_comparison} and Table~\ref{tab:psnr}, none of the baseline methods successfully attack NR-IQA models. 
However, our C-BAIQA even achieves a comparable performance to our P-BAIQA on the LIVEC.
On the KonIQ-10k, we observed a significant drop in the performance compared to P-BAIQA. This may be attributed to the fact that KonIQ-10k contains a larger number of clean data samples compared to LIVEC. Consequently, the NR-IQA models may have been less inclined to learn spurious correlations from the poisoned data and instead relied more on learning from the clean data.
From Fig.~\ref{mra} (b), we see a similar trend as {the results} of P-BAIQA.

\subsection{Resistance to Backdoor Defenses}
We assess the resistance of our attacks to backdoor defenses. Given that strategies like Neural Cleanse~\cite{wang2019neural}, and STRIP~\cite{gao2021design} are tailored for defending against BA in classification, we delve into the resistance of our BAIQA against fine-tuning~\cite{liu2017neural,liu2018fine} and model pruning~\cite{liu2018fine,wu2021adversarial}. 
These are representative defenses that are applicable to our tasks. 
Further details on the setup are in the appendix.

As depicted in Fig.~\ref{resistance}(a), our attacks are resist to fine-tuning. Notably, the mMAE for both P-BAIQA and C-BAIQA remain largely unaffected, experiencing only a marginal increase of approximately 2 after the whole fine-tuning. Furthermore, our attacks are resistant to model pruning, as evident from Fig.~\ref{resistance}(b). Specifically, when applying a pruning rate of 40\%, the mMAE values for P-BAIQA and C-BAIQA remain relatively stable, whereas the benign metric (RMSE) is significantly impacted, resulting in elevated values. 

\subsection{Abaltion Study}
\noindent\textbf{Sampling of $\boldsymbol{\alpha}$ for P-BAIQA.} We explore three cases: 1) {{$\{\pm1,\pm\frac{3}{4},\pm\frac{1}{2},\pm\frac{1}{4}\}$}} with probabilities {{$\{0.4,0.3,0.2,0.1\}$}} as ours; 2) {{$\{\pm1\}$}} with equal probability; 3) Uniform distribution (i.e., {{$\mathrm{U}(-1,1)$}}).
In Table~\ref{table:ablation}, when the quantity of poisoned data is limited, \textit{e.g.,} on LIVEC, option 2) is a preferable choice. 
This preference stems from the fact that a higher absolute value of $\alpha$ results in a greater loss, thereby taking precedence during training, and making the model {more} effectively learn the correlation between the trigger and the attack target.
When the amount of poisoned data is adequate, \textit{e.g.,} on KonIQ-10k, option 1), as adopted in P-BAIQA, provides more precise control regarding the mMAE during attacking.

\noindent\textbf{Value $\boldsymbol{\mu_y}$ and strategy of converting $\boldsymbol{x}$ into $\boldsymbol{x^{\prime}}$ for C-BAIQA.} 
From Table~\ref{table:ablation}, the results of \ding{174} show that the transformation of $\boldsymbol{x}$ into $\boldsymbol{x^{\prime}}$
enhances the attack's efficacy, consistent with Remark 2.
Results of \ding{174} show that values of ${\mu_y}$
that {is} closer to the mean MOS across the dataset {yields} improved attack effectiveness, which is in line with Remark 1.

\section{Conclusion}
In this work, we introduce a novel poisoning-based backdoor attack against NR-IQA, that leverages the adjustment of a scaling coefficient $\alpha$ for the trigger to manipulate the model's output to desired target values. 
We propose embedding the trigger in the DCT domain, and incorporate UAP in the DCT space as the trigger, enhancing the IQA model's susceptibility to manipulation and improving the attack's efficacy. 
Our approach explores both poison-label and clean-label scenarios, the latter focusing on $\alpha$ sampling and image data refinement with theoretical insights.
Extensive experiments on diverse datasets and NR-IQA models validate the effectiveness. 

\section{Acknowledgments}
This work was done at Rapid-Rich Object Search Lab, School of Electrical and Electronic Engineering, Nanyang Technological University. This research is supported in part by the Basic and Frontier Research Project of PCL, the Major Key Project of PCL, and Guangdong Basic and Applied Basic Research Foundation under Grant 2024A1515010454.

\bibliography{aaai25}

\begin{thebibliography}{63}
\providecommand{\natexlab}[1]{#1}

\bibitem[{Chen et~al.(2017)Chen, Liu, Li, Lu, and Song}]{chen2017targeted}
Chen, X.; Liu, C.; Li, B.; Lu, K.; and Song, D. 2017.
\newblock Targeted backdoor attacks on deep learning systems using data poisoning.
\newblock \emph{arXiv preprint arXiv:1712.05526}.

\bibitem[{Cheng et~al.(2023)Cheng, Yang, Zhou, Guo, and Wen}]{cheng2023frequency}
Cheng, H.; Yang, S.; Zhou, J.~T.; Guo, L.; and Wen, B. 2023.
\newblock Frequency guidance matters in few-shot learning.
\newblock In \emph{Proc.~IEEE Int'l Conf.~Computer Vision}, 11814--11824.

\bibitem[{Ding et~al.(2020)Ding, Ma, Wang, and Simoncelli}]{dists}
Ding, K.; Ma, K.; Wang, S.; and Simoncelli, E.~P. 2020.
\newblock Image quality assessment: Unifying structure and texture similarity.
\newblock \emph{{IEEE} Trans. on Pattern Analysis and Machine Intelligence}, 44(5): 2567--2581.

\bibitem[{Doan et~al.(2021)Doan, Lao, Zhao, and Li}]{doan2021lira}
Doan, K.; Lao, Y.; Zhao, W.; and Li, P. 2021.
\newblock LIRA: Learnable, Imperceptible and Robust Backdoor Attacks.
\newblock In \emph{Proc.~IEEE Int'l Conf.~Computer Vision}, 11966--11976.

\bibitem[{Dosovitskiy et~al.(2021)Dosovitskiy, Beyer, Kolesnikov, Weissenborn, Zhai, Unterthiner, Dehghani, Minderer, Heigold, Gelly et~al.}]{dosovitskiy2020image}
Dosovitskiy, A.; Beyer, L.; Kolesnikov, A.; Weissenborn, D.; Zhai, X.; Unterthiner, T.; Dehghani, M.; Minderer, M.; Heigold, G.; Gelly, S.; et~al. 2021.
\newblock An Image is Worth 16x16 Words: Transformers for Image Recognition at Scale.
\newblock In \emph{Proc.~Int'l Conf.~Learning Representations}.

\bibitem[{Dumford and Scheirer(2020)}]{dumford2020backdooring}
Dumford, J.; and Scheirer, W. 2020.
\newblock Backdooring convolutional neural networks via targeted weight perturbations.
\newblock In \emph{Proc. IEEE Int'l Joint Conf. on Biometrics}, 1--9.

\bibitem[{Fang and Choromanska(2022)}]{fang2022backdoor}
Fang, S.; and Choromanska, A. 2022.
\newblock Backdoor attacks on the DNN interpretation system.
\newblock In \emph{Proc.~AAAI Conf. on Artificial Intelligence}, volume~36, 561--570.

\bibitem[{Fu et~al.(2023)Fu, Liang, Liang, Li, Zhang, and Han}]{fu2023asymmetric}
Fu, H.; Liang, F.; Liang, J.; Li, B.; Zhang, G.; and Han, J. 2023.
\newblock Asymmetric learned image compression with multi-scale residual block, importance scaling, and post-quantization filtering.
\newblock \emph{{IEEE} Trans. on Circuits and Systems for Video Technology}.

\bibitem[{Gao et~al.(2021)Gao, Kim, Doan, Zhang, Zhang, Nepal, Ranasinghe, and Kim}]{gao2021design}
Gao, Y.; Kim, Y.; Doan, B.~G.; Zhang, Z.; Zhang, G.; Nepal, S.; Ranasinghe, D.~C.; and Kim, H. 2021.
\newblock Design and evaluation of a multi-domain trojan detection method on deep neural networks.
\newblock \emph{IEEE Trans. on Dependable and Secure Computing}, 19(4): 2349--2364.

\bibitem[{Ghadiyaram and Bovik(2015)}]{ghadiyaram2015massive}
Ghadiyaram, D.; and Bovik, A.~C. 2015.
\newblock Massive online crowdsourced study of subjective and objective picture quality.
\newblock \emph{{IEEE} Trans. on Image Processing}, 25(1): 372--387.

\bibitem[{Ghadiyaram and Bovik(2017)}]{ghadiyaram2017perceptual}
Ghadiyaram, D.; and Bovik, A.~C. 2017.
\newblock Perceptual quality prediction on authentically distorted images using a bag of features approach.
\newblock \emph{Journal of vision}, 17(1): 32--32.

\bibitem[{Golestaneh, Dadsetan, and Kitani(2022)}]{golestaneh2021no}
Golestaneh, S.~A.; Dadsetan, S.; and Kitani, K.~M. 2022.
\newblock No-Reference Image Quality Assessment via Transformers, Relative Ranking, and Self-Consistency.
\newblock In \emph{Proc. of the IEEE/CVF Winter Conf. on Applications of Computer Vision}.

\bibitem[{Gu, Dolan-Gavitt, and Garg(2017)}]{gu2017badnets}
Gu, T.; Dolan-Gavitt, B.; and Garg, S. 2017.
\newblock Badnets: Identifying vulnerabilities in the machine learning model supply chain.
\newblock \emph{arXiv preprint arXiv:1708.06733}.

\bibitem[{Guo, Wu, and Weinberger(2020)}]{guo2020trojannet}
Guo, C.; Wu, R.; and Weinberger, K.~Q. 2020.
\newblock Trojannet: Embedding hidden trojan horse models in neural networks.
\newblock \emph{arXiv preprint arXiv:2002.10078}.

\bibitem[{He et~al.(2016)He, Zhang, Ren, and Sun}]{he2016deep}
He, K.; Zhang, X.; Ren, S.; and Sun, J. 2016.
\newblock Deep residual learning for image recognition.
\newblock In \emph{Proc.~IEEE Int'l Conf.~Computer Vision and Pattern Recognition}, 770--778.

\bibitem[{Hosu et~al.(2020)Hosu, Lin, Sziranyi, and Saupe}]{hosu2020koniq}
Hosu, V.; Lin, H.; Sziranyi, T.; and Saupe, D. 2020.
\newblock KonIQ-10k: An ecologically valid database for deep learning of blind image quality assessment.
\newblock \emph{{IEEE} Trans. on Image Processing}.

\bibitem[{Ilyas et~al.(2018)Ilyas, Engstrom, Athalye, and Lin}]{ilyas2018black}
Ilyas, A.; Engstrom, L.; Athalye, A.; and Lin, J. 2018.
\newblock Black-box adversarial attacks with limited queries and information.
\newblock In \emph{Proc.~Int'l Conf.~Machine Learning}, 2137--2146.

\bibitem[{Ke et~al.(2021)Ke, Wang, Wang, Milanfar, and Yang}]{ke2021musiq}
Ke, J.; Wang, Q.; Wang, Y.; Milanfar, P.; and Yang, F. 2021.
\newblock Musiq: Multi-scale image quality transformer.
\newblock In \emph{Proc.~IEEE Int'l Conf.~Computer Vision}, 5148--5157.

\bibitem[{Korhonen and You(2022)}]{korhonen2022adversarial}
Korhonen, J.; and You, J. 2022.
\newblock Adversarial attacks against blind image quality assessment models.
\newblock In \emph{Proc. of the 2nd Workshop on Quality of Experience in Visual Multimedia Applications}.

\bibitem[{Li et~al.(2020{\natexlab{a}})Li, Xue, Zhao, Zhu, and Zhang}]{li2020invisible}
Li, S.; Xue, M.; Zhao, B. Z.~H.; Zhu, H.; and Zhang, X. 2020{\natexlab{a}}.
\newblock Invisible backdoor attacks on deep neural networks via steganography and regularization.
\newblock \emph{{IEEE} Trans. on Dependable and Secure Computing}, 18(5): 2088--2105.

\bibitem[{Li et~al.(2021)Li, Li, Wu, Li, He, and Lyu}]{li2021invisible}
Li, Y.; Li, Y.; Wu, B.; Li, L.; He, R.; and Lyu, S. 2021.
\newblock Invisible backdoor attack with sample-specific triggers.
\newblock In \emph{Proc.~IEEE Int'l Conf.~Computer Vision}, 16463--16472.

\bibitem[{Li et~al.(2020{\natexlab{b}})Li, Wu, Jiang, Li, and Xia}]{li2020backdoor}
Li, Y.; Wu, B.; Jiang, Y.; Li, Z.; and Xia, S.-T. 2020{\natexlab{b}}.
\newblock Backdoor learning: A survey.
\newblock \emph{arXiv preprint arXiv:2007.08745}.

\bibitem[{Liang et~al.(2021)Liang, Zhang, Wang, Yang, Koyejo, and Li}]{liang2021uncovering}
Liang, K.; Zhang, J.~Y.; Wang, B.; Yang, Z.; Koyejo, S.; and Li, B. 2021.
\newblock Uncovering the connections between adversarial transferability and knowledge transferability.
\newblock In \emph{Proc.~Int'l Conf.~Machine Learning}.

\bibitem[{Liu, Dolan-Gavitt, and Garg(2018)}]{liu2018fine}
Liu, K.; Dolan-Gavitt, B.; and Garg, S. 2018.
\newblock Fine-Pruning: Defending Against Backdooring Attacks on Deep Neural Networks.
\newblock \emph{arXiv preprint arXiv:1805.12185}.

\bibitem[{Liu et~al.(2020)Liu, Ma, Bailey, and Lu}]{liu2020reflection}
Liu, Y.; Ma, X.; Bailey, J.; and Lu, F. 2020.
\newblock Reflection backdoor: A natural backdoor attack on deep neural networks.
\newblock In \emph{European Conf. on Computer Vision}, 182--199.

\bibitem[{Liu, Xie, and Srivastava(2017)}]{liu2017neural}
Liu, Y.; Xie, Y.; and Srivastava, A. 2017.
\newblock Neural trojans.
\newblock In \emph{IEEE Int'l Conf. on Computer Design}, 45--48.

\bibitem[{Liu et~al.(2024{\natexlab{a}})Liu, Yang, Li, Ding, and Jiang}]{liu2024defense}
Liu, Y.; Yang, C.; Li, D.; Ding, J.; and Jiang, T. 2024{\natexlab{a}}.
\newblock Defense Against Adversarial Attacks on No-Reference Image Quality Models with Gradient Norm Regularization.
\newblock \emph{arXiv preprint arXiv:2403.11397}.

\bibitem[{Liu et~al.(2024{\natexlab{b}})Liu, Wang, Huai, and Miao}]{liu2024backdoor}
Liu, Z.; Wang, T.; Huai, M.; and Miao, C. 2024{\natexlab{b}}.
\newblock Backdoor attacks via machine unlearning.
\newblock In \emph{Proc.~AAAI Conf. on Artificial Intelligence}, volume~38, 14115--14123.

\bibitem[{Madry et~al.(2018)Madry, Makelov, Schmidt, Tsipras, and Vladu}]{DBLP:conf/iclr/MadryMSTV18}
Madry, A.; Makelov, A.; Schmidt, L.; Tsipras, D.; and Vladu, A. 2018.
\newblock Towards Deep Learning Models Resistant to Adversarial Attacks.
\newblock In \emph{Proc.~Int'l Conf.~Learning Representations}.

\bibitem[{Mittal, Moorthy, and Bovik(2012)}]{mittal2012no}
Mittal, A.; Moorthy, A.~K.; and Bovik, A.~C. 2012.
\newblock No-reference image quality assessment in the spatial domain.
\newblock \emph{{IEEE} Trans. on Image Processing}, 21(12): 4695--4708.

\bibitem[{Nguyen and Tran(2020)}]{nguyen2020input}
Nguyen, T.~A.; and Tran, A. 2020.
\newblock Input-aware dynamic backdoor attack.
\newblock In \emph{Proc.~Annual Conf.~Neural Information Processing Systems}, volume~33, 3454--3464.

\bibitem[{Nguyen and Tran(2021)}]{nguyen2021wanet}
Nguyen, T.~A.; and Tran, A.~T. 2021.
\newblock WaNet - Imperceptible Warping-based Backdoor Attack.
\newblock In \emph{Proc.~Int'l Conf.~Learning Representations}.

\bibitem[{Rakin, He, and Fan(2020)}]{rakin2020tbt}
Rakin, A.~S.; He, Z.; and Fan, D. 2020.
\newblock Tbt: Targeted neural network attack with bit trojan.
\newblock In \emph{Proc.~IEEE Int'l Conf.~Computer Vision and Pattern Recognition}, 13198--13207.

\bibitem[{Rippel et~al.(2019)Rippel, Nair, Lew, Branson, Anderson, and Bourdev}]{rippel2019learned}
Rippel, O.; Nair, S.; Lew, C.; Branson, S.; Anderson, A.~G.; and Bourdev, L. 2019.
\newblock Learned video compression.
\newblock In \emph{Proc.~IEEE Int'l Conf.~Computer Vision}, 3454--3463.

\bibitem[{Saha, Subramanya, and Pirsiavash(2020)}]{saha2020hidden}
Saha, A.; Subramanya, A.; and Pirsiavash, H. 2020.
\newblock Hidden trigger backdoor attacks.
\newblock In \emph{Proc.~AAAI Conf. on Artificial Intelligence}, volume~34, 11957--11965.

\bibitem[{Shumitskaya, Antsiferova, and Vatolin(2022)}]{shumitskaya2022universal}
Shumitskaya, E.; Antsiferova, A.; and Vatolin, D. 2022.
\newblock Universal perturbation attack on differentiable no-reference image-and video-quality metrics.
\newblock In \emph{BMVC}.

\bibitem[{Steinhardt, Koh, and Liang(2017)}]{steinhardt2017certified}
Steinhardt, J.; Koh, P. W.~W.; and Liang, P.~S. 2017.
\newblock Certified defenses for data poisoning attacks.
\newblock In \emph{Proc.~Annual Conf.~Neural Information Processing Systems}, volume~30.

\bibitem[{Su et~al.(2020)Su, Yan, Zhu, Zhang, Ge, Sun, and Zhang}]{su2020blindly}
Su, S.; Yan, Q.; Zhu, Y.; Zhang, C.; Ge, X.; Sun, J.; and Zhang, Y. 2020.
\newblock Blindly assess image quality in the wild guided by a self-adaptive hyper network.
\newblock In \emph{Proc.~IEEE Int'l Conf.~Computer Vision and Pattern Recognition}, 3667--3676.

\bibitem[{Szegedy et~al.(2013)Szegedy, Zaremba, Sutskever, Bruna, Erhan, Goodfellow, and Fergus}]{intriguing}
Szegedy, C.; Zaremba, W.; Sutskever, I.; Bruna, J.; Erhan, D.; Goodfellow, I.; and Fergus, R. 2013.
\newblock Intriguing properties of neural networks.
\newblock \emph{arXiv preprint arXiv:1312.6199}.

\bibitem[{Wallace(1992)}]{wallace1992jpeg}
Wallace, G.~K. 1992.
\newblock The JPEG still picture compression standard.
\newblock \emph{{IEEE} Trans. on Consumer Electronics}, 38: 43--59.

\bibitem[{Wang et~al.(2019)Wang, Yao, Shan, Li, Viswanath, Zheng, and Zhao}]{wang2019neural}
Wang, B.; Yao, Y.; Shan, S.; Li, H.; Viswanath, B.; Zheng, H.; and Zhao, B.~Y. 2019.
\newblock Neural cleanse: Identifying and mitigating backdoor attacks in neural networks.
\newblock In \emph{IEEE Symposium on Security and Privacy}, 707--723.

\bibitem[{Wang et~al.(2022)Wang, Yao, Xu, An, Tong, and Wang}]{wang2021backdoor}
Wang, T.; Yao, Y.; Xu, F.; An, S.; Tong, H.; and Wang, T. 2022.
\newblock An invisible black-box backdoor attack through frequency domain.
\newblock In \emph{European Conf. on Computer Vision}.

\bibitem[{Wang et~al.(2024{\natexlab{a}})Wang, Hu, Zhang, Zhou, Zhang, Xu, Wan, and Jin}]{wang2024eclipse}
Wang, X.; Hu, S.; Zhang, Y.; Zhou, Z.; Zhang, L.~Y.; Xu, P.; Wan, W.; and Jin, H. 2024{\natexlab{a}}.
\newblock ECLIPSE: Expunging clean-label indiscriminate poisons via sparse diffusion purification.
\newblock In \emph{European Symposium on Research in Computer Security}.

\bibitem[{Wang et~al.(2024{\natexlab{b}})Wang, Li, Liu, Zhang, Hu, Zhang, Zhou, and Jin}]{wang2024unlearnable}
Wang, X.; Li, M.; Liu, W.; Zhang, H.; Hu, S.; Zhang, Y.; Zhou, Z.; and Jin, H. 2024{\natexlab{b}}.
\newblock Unlearnable 3{D} point clouds: Class-wise transformation is all you need.
\newblock In \emph{Proc.~Annual Conf.~Neural Information Processing Systems}.

\bibitem[{Wang et~al.(2004)Wang, Bovik, Sheikh, and Simoncelli}]{ssim}
Wang, Z.; Bovik, A.~C.; Sheikh, H.~R.; and Simoncelli, E.~P. 2004.
\newblock Image quality assessment: from error visibility to structural similarity.
\newblock \emph{{IEEE} Trans. on Image Processing}.

\bibitem[{Wu and Wang(2021)}]{wu2021adversarial}
Wu, D.; and Wang, Y. 2021.
\newblock Adversarial neuron pruning purifies backdoored deep models.
\newblock In \emph{Advances in Neural Information Processing Systems}, volume~34, 16913--16925.

\bibitem[{Xia et~al.(2024{\natexlab{a}})Xia, Yang, Yu, Lin, Ding, DUAN, and Jiang}]{xiatransferable}
Xia, S.; Yang, W.; Yu, Y.; Lin, X.; Ding, H.; DUAN, L.; and Jiang, X. 2024{\natexlab{a}}.
\newblock Transferable Adversarial Attacks on SAM and Its Downstream Models.
\newblock In \emph{Proc.~Annual Conf.~Neural Information Processing Systems}.

\bibitem[{Xia et~al.(2024{\natexlab{b}})Xia, Yu, Jiang, and Ding}]{xia2024mitigating}
Xia, S.; Yu, Y.; Jiang, X.; and Ding, H. 2024{\natexlab{b}}.
\newblock Mitigating the Curse of Dimensionality for Certified Robustness via Dual Randomized Smoothing.
\newblock In \emph{Proc.~Int'l Conf.~Learning Representations}.

\bibitem[{Yang et~al.(2022)Yang, Wu, Shi, Lao, Gong, Cao, Wang, and Yang}]{yang2022maniqa}
Yang, S.; Wu, T.; Shi, S.; Lao, S.; Gong, Y.; Cao, M.; Wang, J.; and Yang, Y. 2022.
\newblock MANIQA: Multi-dimension Attention Network for No-Reference Image Quality Assessment.
\newblock \emph{arXiv preprint arXiv:2204.08958}.

\bibitem[{Yu et~al.(2024{\natexlab{a}})Yu, Zeng, Zhao, Pang, and Wang}]{yu2024chronic}
Yu, F.; Zeng, B.; Zhao, K.; Pang, Z.; and Wang, L. 2024{\natexlab{a}}.
\newblock Chronic Poisoning: Backdoor Attack against Split Learning.
\newblock In \emph{Proc.~AAAI Conf. on Artificial Intelligence}.

\bibitem[{Yu et~al.(2024{\natexlab{b}})Yu, Wang, Xia, Yang, Lu, Tan, and Kot}]{yu2024purify}
Yu, Y.; Wang, Y.; Xia, S.; Yang, W.; Lu, S.; Tan, Y.-P.; and Kot, A.~C. 2024{\natexlab{b}}.
\newblock Purify Unlearnable Examples via Rate-Constrained Variational Autoencoders.
\newblock In \emph{Proc.~Int'l Conf.~Machine Learning}.

\bibitem[{Yu et~al.(2024{\natexlab{c}})Yu, Wang, Yang, Guo, Lu, Duan, Tan, and Kot}]{yu2024robust}
Yu, Y.; Wang, Y.; Yang, W.; Guo, L.; Lu, S.; Duan, L.-Y.; Tan, Y.-P.; and Kot, A.~C. 2024{\natexlab{c}}.
\newblock Robust and Transferable Backdoor Attacks Against Deep Image Compression With Selective Frequency Prior.
\newblock \emph{{IEEE} Trans. on Pattern Analysis and Machine Intelligence}.

\bibitem[{Yu et~al.(2023)Yu, Wang, Yang, Lu, Tan, and Kot}]{yu2023backdoor}
Yu, Y.; Wang, Y.; Yang, W.; Lu, S.; Tan, Y.-P.; and Kot, A.~C. 2023.
\newblock Backdoor attacks against deep image compression via adaptive frequency trigger.
\newblock In \emph{Proc.~IEEE Int'l Conf.~Computer Vision and Pattern Recognition}, 12250--12259.

\bibitem[{Yu et~al.(2022)Yu, Yang, Tan, and Kot}]{Yu_2022_CVPR}
Yu, Y.; Yang, W.; Tan, Y.-P.; and Kot, A.~C. 2022.
\newblock Towards Robust Rain Removal Against Adversarial Attacks: A Comprehensive Benchmark Analysis and Beyond.
\newblock In \emph{Proc.~IEEE Int'l Conf.~Computer Vision and Pattern Recognition}.

\bibitem[{Yue et~al.(2022)Yue, Lv, Liang, and Chen}]{yue2022invisible}
Yue, C.; Lv, P.; Liang, R.; and Chen, K. 2022.
\newblock Invisible backdoor attacks using data poisoning in the frequency domain.
\newblock \emph{arXiv preprint arXiv:2207.04209}.

\bibitem[{Zeng et~al.(2021)Zeng, Park, Mao, and Jia}]{zeng2021rethinking}
Zeng, Y.; Park, W.; Mao, Z.~M.; and Jia, R. 2021.
\newblock Rethinking the backdoor attacks' triggers: A frequency perspective.
\newblock In \emph{Proc.~IEEE Int'l Conf.~Computer Vision}, 16473--16481.

\bibitem[{Zhang et~al.(2023)Zhang, Ran, Tang, and Wang}]{zhang2024vulnerabilities}
Zhang, A.; Ran, Y.; Tang, W.; and Wang, Y.-G. 2023.
\newblock Vulnerabilities in video quality assessment models: The challenge of adversarial attacks.
\newblock In \emph{Proc.~Annual Conf.~Neural Information Processing Systems}, volume~36.

\bibitem[{Zhang et~al.(2018)Zhang, Isola, Efros, Shechtman, and Wang}]{lpips}
Zhang, R.; Isola, P.; Efros, A.~A.; Shechtman, E.; and Wang, O. 2018.
\newblock The unreasonable effectiveness of deep features as a perceptual metric.
\newblock In \emph{Proc.~IEEE Int'l Conf.~Computer Vision and Pattern Recognition}, 586--595.

\bibitem[{Zhang et~al.(2022)Zhang, Li, Min, Zhai, Guo, Yang, and Ma}]{zhang2022perceptual}
Zhang, W.; Li, D.; Min, X.; Zhai, G.; Guo, G.; Yang, X.; and Ma, K. 2022.
\newblock Perceptual attacks of no-reference image quality models with human-in-the-loop.
\newblock In \emph{Proc.~Annual Conf.~Neural Information Processing Systems}.

\bibitem[{Zhang et~al.(2019)Zhang, Liu, Dong, and Qiao}]{zhang2019ranksrgan}
Zhang, W.; Liu, Y.; Dong, C.; and Qiao, Y. 2019.
\newblock Ranksrgan: Generative adversarial networks with ranker for image super-resolution.
\newblock In \emph{Proc.~IEEE Int'l Conf.~Computer Vision}.

\bibitem[{Zhang et~al.(2020)Zhang, Ma, Yan, Deng, and Wang}]{zhang2020blind}
Zhang, W.; Ma, K.; Yan, J.; Deng, D.; and Wang, Z. 2020.
\newblock Blind Image Quality Assessment Using A Deep Bilinear Convolutional Neural Network.
\newblock \emph{{IEEE} Trans. on Circuits and Systems for Video Technology}, 30(1): 36--47.

\bibitem[{Zhang et~al.(2021)Zhang, Ma, Zhai, and Yang}]{zhang2021uncertainty}
Zhang, W.; Ma, K.; Zhai, G.; and Yang, X. 2021.
\newblock Uncertainty-aware blind image quality assessment in the laboratory and wild.
\newblock \emph{{IEEE} Trans. on Image Processing}, 30: 3474--3486.

\bibitem[{Zheng et~al.(2024)Zheng, Yu, Yang, Liu, Lam, and Kot}]{zheng2024towards}
Zheng, Q.; Yu, Y.; Yang, S.; Liu, J.; Lam, K.-Y.; and Kot, A. 2024.
\newblock Towards Physical World Backdoor Attacks against Skeleton Action Recognition.
\newblock In \emph{European Conf. on Computer Vision}.

\end{thebibliography}

\newpage
\clearpage

\appendix
\section{Appendix}
\subsection{Impact Statement}
\label{appendix:impact_statement}
The research presented in this paper, focusing on backdoor attacks against No-Reference Image Quality Assessment (NR-IQA) models, while valuable for understanding and defending against potential threats, also brings forth significant social implications. NR-IQA systems are crucial for a wide range of applications, including but not limited to medical imaging, autonomous driving, and surveillance systems. As such, vulnerabilities in these systems have the potential to cause harm at both individual and societal levels.
Firstly, successful backdoor attacks on NR-IQA models can be leveraged for malicious purposes, such as altering the perception of image quality in critical applications. In medical imaging, this could lead to misdiagnosis or incorrect treatment plans. In autonomous driving, it could result in the system misinterpreting road conditions, potentially leading to accidents. In surveillance systems, attackers could potentially manipulate video feeds to evade detection.
Secondly, the methods presented in this research demonstrate that current NR-IQA models are susceptible to manipulation, highlighting the need for greater attention and resources towards enhancing their security and robustness. As NR-IQA systems become more ubiquitous in society, their potential impact also increases, and it is essential that researchers and developers prioritize security considerations in their design and implementation.
Finally, this research also raises awareness of the broader issue of adversarial machine learning, and the need for continued research in this area. As machine learning and AI technologies continue to proliferate, the potential for malicious use of these technologies also increases. By studying and defending against adversarial attacks, we can help ensure that these technologies are used responsibly and safely.
In summary, while the research presented in this abstract provides valuable insights into vulnerabilities in NR-IQA models, it also highlights the need for greater attention towards the social implications of these vulnerabilities, and the need for continued research in adversarial machine learning to ensure the safe and responsible use of AI technologies.


\subsection{Formulations of RMSE, SROCC, and PLCC}
\label{appendix:formulations}
In this section, we will introduce IQA-specific metrics RMSE, SROCC, and PLCC in Sec~\textcolor{red}{4.1}.

\textbf{RMSE} measures the difference between MOS values and predicted scores, which is represented as
\begin{equation}
    \text{RMSE} = \sqrt{\frac{1}{N}
    \sum_{i=1}^N (y_i-f_i)^2}.
\end{equation}
In this equation, $N$ is the number of images. $y_i$ and $f_i$ represent the MOS and predicted score of the $i^{th}$ image, respectively. The smaller the RMSE value is, the smaller the differences between the two groups of scores.

\textbf{SROCC} measures the correlation between MOS values and predicted scores to what extent the correlation can be described by a monotone function. The specific formulation is as follows:
\begin{equation}
    \text{SROCC} = 1 - \frac{6 \sum_{i=1}^N d_i^2}{ N(N^2-1)},
\end{equation}
where $d_i$ denotes the difference between orders of the $i^{th}$ image in subjective and objective quality scores. The closer the SROCC value is to 1, the more consistent the ordering is between two groups of scores. 

\textbf{PLCC} measures the linear correlation between MOS values and predicted scores, which is formulated as
\begin{equation}
\begin{split}
    \text{PLCC} &= \frac{
    \sum_{i=1}^N (y_i-\Bar{y})(f_i - \Bar{f})
    }{
    \sum_{i=1}^N (y_i-\Bar{y})^2(f_i - \Bar{f})^2
    }, \\
    \Bar{y} & = \frac{1}{N} \sum_{i=1}^N y_i,  \Bar{f} = \frac{1}{N} \sum_{i=1}^N f_i.
\end{split}
\end{equation}
The closer the PLCC value is to 1, the higher the positive correlation between the two groups of scores.

\subsection{Hardware Setup}
\label{appendix:harware_setup}
We conducted all the training, test, and attack on an NVIDIA GeForce RTX 3090 GPU with 24GB of memory.

\subsection{Standard Training of NR-IQA models}
\label{appendix:training}
For the standard training of three NR-IQA models: HyperIQA~\cite{su2020blindly}, DBCNN~\cite{zhang2020blind}, and TReS~\cite{golestaneh2021no}, we follow their official code, and model architecture, \textit{i.e.,} ResNet-50 as the feature extractor for HyperIQA, VGG for DBCNN, and ResNet-18 for TReS.

When considering data augmentations and evaluation for testing, we establish a cropping size of $224\times224$ and utilize 25 patches per image.
For all models, a batch size of 32 is adopted.

\noindent\textbf{HyperIQA.} Specifically for the HyperIQA model, we select the L1 loss function, employ the Adam optimizer, set an initial learning rate of 2e-5, a weight decay of 5e-4, and train for 24 epochs. Additionally, we incorporate a multistep learning rate scheduler with a step of 8 epochs and a gamma value of 0.1.

\noindent\textbf{DBCNN.} For the DBCNN model, we choose the MSE loss. The initial 8 epochs are dedicated to training the fully-connected layer alone, utilizing the SGD optimizer with a fixed learning rate of 1e-3 and a weight decay of 5e-4. In the subsequent 8 epochs, we switch to the Adam optimizer, maintaining a fixed learning rate of 1e-5 and the same weight decay.

\noindent\textbf{TReS.} Lastly, for the TReS model, we opt for the L1 loss, utilize the Adam optimizer, set an initial learning rate of 2e-5, a weight decay of 5e-4, and train for 12 epochs. A multistep learning rate scheduler with a step of 4 epochs and a gamma value of 0.1 is also employed.

\subsection{Detailed Setup of The Baseline Attacks}
\label{appendix:attacks}
We have incorporated Blended~\cite{chen2017targeted} and WaNet~\cite{nguyen2021wanet} into our work by utilizing the open-source Python toolbox, BackdoorBox~\cite{li2023backdoorbox}.
For Blended, when testing on the LIVEC dataset, we randomly sample the noise from a uniform distribution $U[0,1]$ and set the weight to 0.1 when the scaling coefficient $\alpha=1$.
Similarly, for WaNet on the LIVEC dataset, we adhere to the default settings, setting the uniform grid size to 4, and the strength to 0.4 when $\alpha=1$.
Furthermore, we have conducted experiments with FTrojan~\cite{wang2021backdoor} using the official codebase. In our experiments, we select the middle 64 frequencies to embed the trigger, and on the LIVEC dataset, we set the magnitude of the trigger to 66 when $\alpha=1$, which is well-aligned with our approach.
When adjusting the strength of the trigger, we apply the $\alpha$ to the weight in Blended, the strength in WaNet, and the magnitude in FTrojan. This allows us to control the intensity or visibility of the trigger in each respective method.
On the Koniq-10k dataset, for a fair comparison with our method, we have reduced the values of all strength-related hyperparameters for the baseline methods by half.

\subsection{Detailed Setup of The Backdoor Defenses}
\label{appendix:defenses}
\noindent\textbf{Settings for Fine-tuning.} As an example for discussion, we perform experiments on the LIVEC dataset. We choose to fine-tune the whole network on a randomly selected benign subset. Specifically, for fine-tuning, we utilize 20\% of the benign training samples, while maintaining all other training configurations in line with the standard training procedures for NR-IQA models.

\noindent\textbf{Settings for Model Pruning.}
We conduct the experiments on the LIVEC dataset as an example for discussion. Following its default settings, we conduct channel pruning~\cite{he2017channel} on the output of the last convolutional layer of the feature extractor with 20\% benign training samples. The pruning rate $\beta \in \{0\%, 5\%,\cdots,95\%\}$.

\begin{table*}[t]
\caption{Evaluation of P-BAIQA with different poisoning ratios.}
\centering
\scalebox{0.69}{
\setlength{\tabcolsep}{2.5pt}
\begin{tabular}{c cc||*3{c}|*3{c}||*3{c}|*3{c}}
\hline
\multirow{3}{*}{\shortstack{Attack\\Type}}&\multicolumn{2}{c||}{Dataset $\rightarrow$} &
\multicolumn{6}{c||}{LIVEC} &
\multicolumn{6}{c}{KonIQ-10k}\\
\cline{2-15}
&\multirow{2}{*}{Model$\downarrow$}& \multirow{2}{*}{Ratio $r$ (\%)$\downarrow$}&\multicolumn{3}{c|}{Benign Metrics}&\multicolumn{3}{c||}{Attack Metrics}&\multicolumn{3}{c|}{Benign Metrics}&\multicolumn{3}{c}{Attack Metrics}
\\
&&& PLCC& SROCC& RMSE &$\text{PSNR}_{1}$& mMAE& mMRA& PLCC& SROCC& RMSE& $\text{PSNR}_{1}$ & mMAE& mMRA\\
\cline{1-15}
\multirow{6}{*}{\rotatebox{90}{P-BAIQA}}&\multirow{2}{*}{{HyperIQA}}& 5 & 0.9052 & 0.8839 & 9.580 & 30.06 & 15.023 & 0.2799 & 0.9011 & 0.8882  & 6.902 & 36.32 & 11.277 & 0.4021\\
&&10 &  0.9051 & 0.8921 & 9.705 & 30.06 & 13.896 & 0.3336 & 0.9071 & 0.8871  & 6.883 & 36.32 & 8.415 & 0.7494\\
\cline{2-15}
&\multirow{2}{*}{{DBCNN}} &5 & 0.8762 & 0.8542 & 10.348 & 30.06 & 13.271 & 0.3517 & 0.9005 & 0.8829  & 7.070 &  36.32& 7.777 & 0.5846\\
&&10 &  0.8667 & 0.8624 & 10.573 & 30.06 & 11.297 & 0.4798 & 0.9009 & 0.8852  & 7.008 & 36.32 & 6.986 & 0.6077\\
\cline{2-15}
&\multirow{2}{*}{{TReS}}& 5 & 0.8825 & 0.8543 & 17.879 & 30.06 & 19.978 & 0.0825 & 0.8945 & 0.8780  & 20.874 & 36.32 & 11.494 & 0.5029\\
&&10 &  0.8788 & 0.8574 & 16.941 & 30.06 & 14.619 & 0.3327 & 0.8994 & 0.8785  & 22.315 & 36.32 & 9.797 & 0.6772\\
\hline
\end{tabular}
}
\label{table:various_ratios}
\end{table*}

\subsection{More Details of Experimental Results}
\label{appendix:experiments}
\noindent\textbf{Main experiments.} Herein, we present an extended analysis and results of the $\text{MAE}(\alpha)$ for both P-BAIQA and C-BAIQA.  
  
\begin{figure}[t]
\centering
\includegraphics[width=1.0\linewidth]{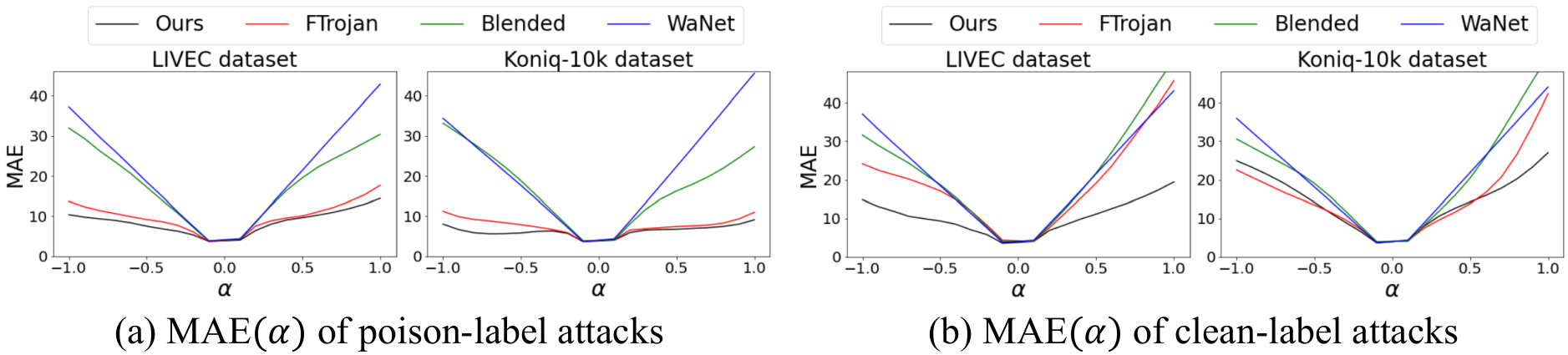}
\caption{
$\text{MAE}(\alpha)$ for both P-BAIQA and C-BAIQA (HyperIQA as victim models).
}
\label{mae}
\end{figure}

As depicted in Fig.~\ref{mra} and \ref{mae}, during clean label attacks, all baseline methods are unable to effectively compromise the NR-IQA models. This deficiency may be attributed to the limited representation capacity of the trigger. Notably, in the case of the Blended approach, it becomes apparent that the backdoored model primarily learns to diminish the output MOS scores regardless of the value of $\alpha$.  
  
Conversely, for poison label attacks, our proposed method significantly outperforms Blended and WaNet, and exhibits slightly superior performance compared to FTrojan.

In addition to mMAE and mMRA, which assess the attack's mean effectiveness, we provide the $\text{MRA}(\alpha)$ in Fig.~\ref{mra} (a). We can see that our method is capable of achieving a significant deviation when $\alpha$ has an absolute value greater than 0.5. However, when $\alpha$ {falls} within [-0.5, 0.5], the manipulation becomes less precise, as the model finds it more challenging to recognize the trigger.
However, we believe this limitation is tolerable given that attackers generally aim for significant shifts in outputs rather than minor alterations.
Results of $\text{MAE}(\alpha)$ and \textbf{visualized results} of poisoned images are in Appendix~\ref{appendix:experiments}.
Moreover, we provide the results of P-BAIQA with lower poisoning ratios in Table~\ref{table:various_ratios} in the appendix, and can observe that higher poisoning ratios indicate better attack effectiveness.

\begin{figure}[t]
\centering
\includegraphics[width=0.8\linewidth]{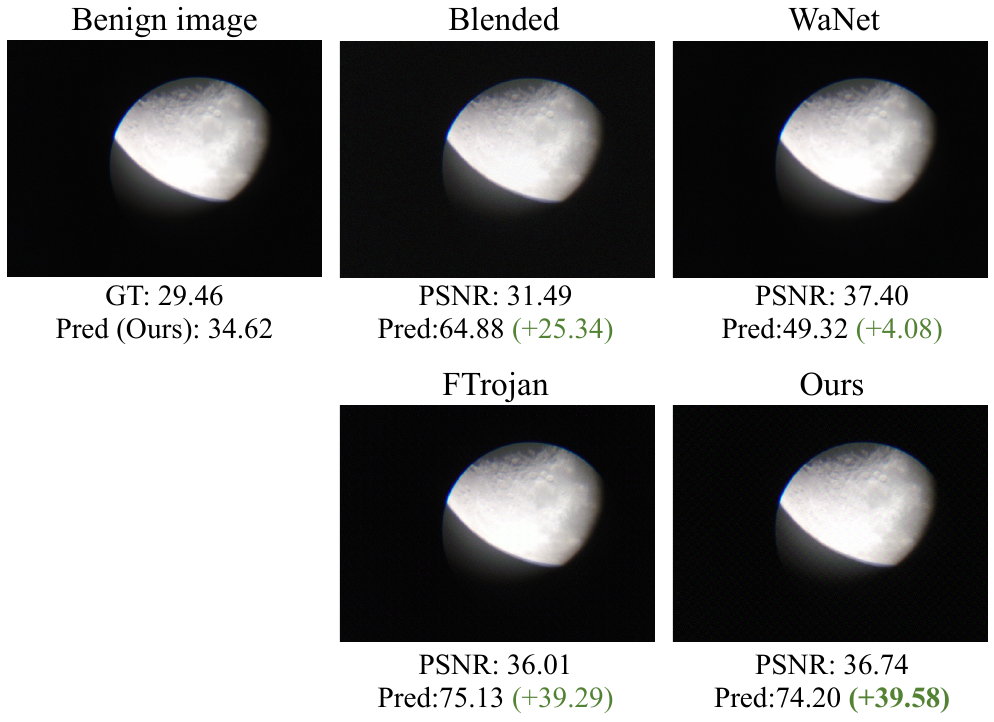}
\caption{
Visualized results of poison-label attacks on Koniq-10k. For all poisoned images, we adopt $\alpha=1$, and the deviations of the output are compared to the clean output of corresponding models.
}
\label{v1}
\end{figure}

\begin{figure}[t]
\centering
\includegraphics[width=0.8\linewidth]{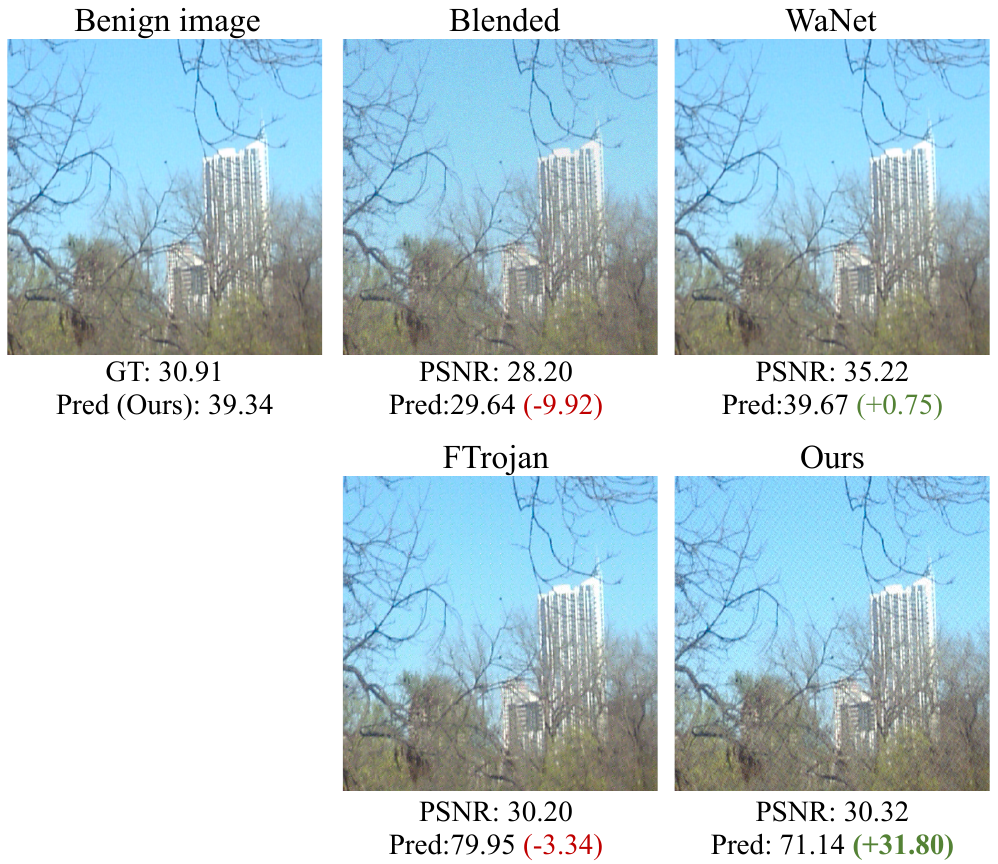}
\caption{
Visualized results of clean-label attacks on LIVEC. For all poisoned images, we adopt $\alpha=1$, and the deviations of the output are compared to the clean output of corresponding models.
}
\label{v2}
\end{figure}

\noindent\textbf{More visualized results.}
In Fig.~\ref{v1} and \ref{v2}, we offer visualizations that illustrate the comparative results of our method and the baseline approaches (we set $\alpha=1$ for poisoned images, and the ideal deviations $\alpha\cdot\Delta{y_t}=40$.). For poison-label attacks, it is evident that our method outperforms the others, demonstrating superior attack performance while minimizing the impact on clean data. When considering clean-label attacks, it becomes apparent that the baseline methods are unable to effectively attack NR-IQA models. Specifically, Blended and FTrojan techniques struggle to achieve significant decreases in output scores, while WaNet exhibits almost no deviation in its performance.

\noindent\textbf{Ablation study.} 
Here, we present the $\text{MRA}(\alpha)$ and $\text{MSE}(\alpha)$ values for P-BAIQA and C-BAIQA, corresponding to the ablation study conducted in Section 5.1.  
  
\begin{figure}[t]
\centering
\includegraphics[width=1.0\linewidth]{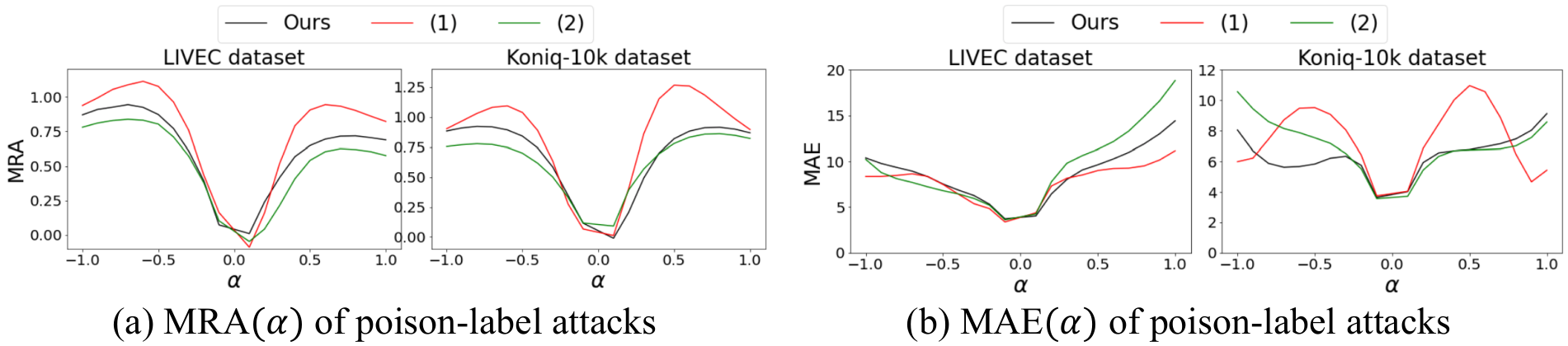}
\caption{
Ablation study: $\text{MRA}(\alpha)$ and $\text{MSE}(\alpha)$ for P-BAIQA (HyperIQA as victim models).
}
\label{ablation1}
\end{figure}

As shown in Fig.~\ref{ablation1}, on both datasets, while approach (1) introduces greater deviations in the output, its manipulation is not precise, often resulting in a higher MAE. In comparison, approach (2) performs worse in terms of both MRA and MAE.  
  
\begin{figure}[t]
\centering
\includegraphics[width=1.0\linewidth]{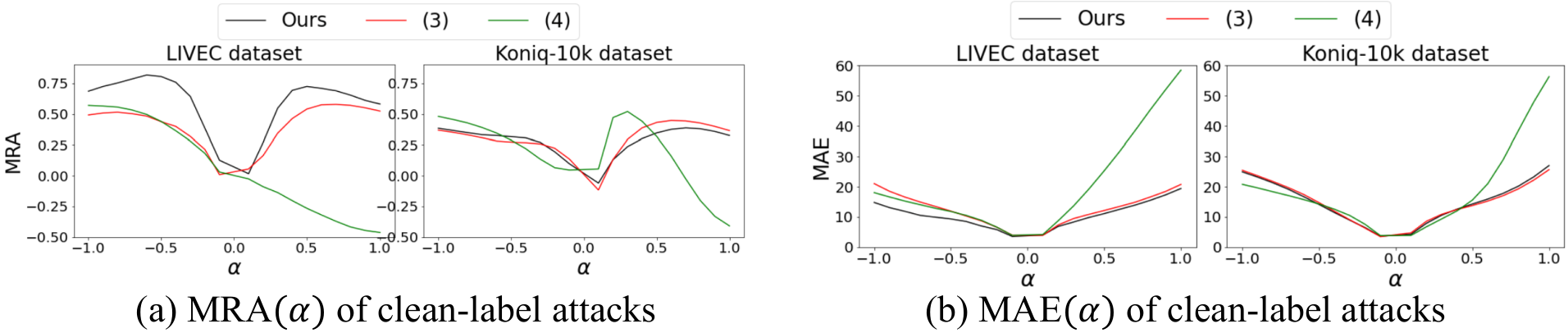}
\caption{
Ablation study: $\text{MRA}(\alpha)$ and $\text{MSE}(\alpha)$ for C-BAIQA (HyperIQA as victim models).
}
\label{ablation2}
\end{figure}

Furthermore, Fig.~\ref{ablation2} demonstrates that converting $\boldsymbol{x}$ to $\boldsymbol{x'}$ effectively enhances the attack performance, as compared to approach (3). The results of approach (4) highlight the crucial role of the $\mu_y$ selection. Specifically, when $\mu_y$ is set to 65, the backdoored model learns to solely decrease the output MOS scores, regardless of the $\alpha$ value chosen for the trigger.


\end{document}